%% file: main.tex
\documentclass[utf8]{frontiersSCNS} 

\usepackage{url,hyperref,microtype,subcaption}
\usepackage[onehalfspacing]{setspace}
\usepackage{bm}
\usepackage{algorithm}
\usepackage{algorithmic}

\def\keyFont{\fontsize{8}{11}\helveticabold }
\def\firstAuthorLast{Hieida {et~al.}} 
\def\Authors{Chie Hieida\,$^{1,*}$, Takato Horii\,$^{1}$ and Takayuki Nagai\,$^{1,2}$}
\def\Address{$^{1}$ Graduate School of Informatics and Engineering, The University of Electro-Communications, Tokyo, Japan \\
$^{2}$ Artificial Intelligence Exploration Research Center, The University of Electro-Communications, Tokyo, Japan }
\def\corrAuthor{Chie Hieida}

\def\corrEmail{hchie@apple.ee.uec.ac.jp}

\begin{document}
\onecolumn
\firstpage{1}

\title[Deep Emotion]{Deep Emotion: A Computational Model of Emotion Using Deep Neural Networks} 

\author[\firstAuthorLast ]{\Authors} 
\address{} 
\correspondance{} 

\extraAuth{}

\maketitle

\begin{abstract}

\section{}
Emotions are very important for human intelligence. 
For example, emotions are closely related to the appraisal of the internal bodily state and external stimuli.  
This helps us to respond quickly to the environment. 
Another important perspective in human intelligence is the role of emotions in decision-making.
Moreover, the social aspect of emotions is also very important. 
Therefore, if the mechanism of emotions were elucidated, we could advance toward the essential understanding of our natural intelligence.  
In this study, a model of emotions is proposed to elucidate the mechanism of emotions through the computational model. 
Furthermore, from the viewpoint of partner robots, the model of emotions may help us to build robots that can have empathy for humans. 
To understand and sympathize with people's feelings, the robots need to have their own emotions. 
This may allow robots to be accepted in human society. 
The proposed model is implemented using deep neural networks consisting of three modules, which interact with each other. 
Simulation results reveal that the proposed model exhibits reasonable behavior as the basic mechanism of emotion. 

%
%
%
\tiny
 \keyFont{ \section{Keywords:} Emotion Model, Human--Robot Interaction, Empathic Communication, Machine Learning, Recurrent Attention Model, Convolutional Long Short-Term Memory, Deep Deterministic Policy Gradient} 
\end{abstract}

\input{intro}

\input{model}

\input{implement}

\input{experiment}
\input{conclusion}

\section*{Author Contributions}

CH and TN conceived of the presented idea. 
CH developed the theory and implemented the system.
CH and TN analyzed the results, and all authors discussed the results.
CH wrote the manuscript with support from TH and TN.


\section*{Acknowledgments}
This research was subsidized by JSPS Science Research Fund JP 16 J 04930, JST CREST (JPMJCR15E3), and Grant-in-Aid for Scientific Research on Innovative Areas (26118001).



\bibliographystyle{frontiersinSCNS_ENG_HUMS} 

\input{main.bbl}




\input{appendix}
\end{document}

%% file: intro.tex
\section{Introduction}
The development of artificial intelligence (AI) in recent years has been remarkable. 
In certain tasks such as object recognition, it is said that AI has surpassed human capabilities. 
However, one might think that emotion separates human intelligence from AI. 
Is this true? 
If the human mind is created as a result of the calculations of the brain, then emotions could be simulated by a computer. 
Would this imply the possibility that a robot could have emotions? 
To answer this question, we should start thinking of the basic mechanism of emotion. 
If the mechanism of emotion were elucidated, we could get closer to the essential understanding of what a human being is. 

Because emotions are very important to human beings, many studies on emotions have been carried out in the past. 
Here, the conventional studies on emotions are organized from the viewpoint of their research approach.
Many psychological studies, among others, have tried to capture emotional phenomena.  
Emotional facial expression studies, for instance, are based on the idea of basic emotions theory. 
It is well known that Ekman insisted that there are six basic emotions, regardless of culture \citep{ekman}. 
Plutchik and Izard respectively assumed eight and ten basic emotions \citep{plutchik1980emotion,Plutchik,izard1977}. 
Because the basic emotions theory is based on the evolutionary point of view, emotional expressions are defined at the nerve level, and the categories of expression and recognition of basic facial expressions are universal. 
The dimensional model of emotions is another well known approach for emotions \citep{Russell,Schlosberg1954}. 
This expresses emotions in approximately two to three cognitive dimensions based on a factor analysis of judgment on emotional stimuli such as expressive photographs or emotional expression words. 
Although many emotion-related studies are based on dimensional models, the mechanism behind emotional phenomena cannot be revealed. 

Various emotional models have been proposed in the literature from the physiological viewpoint \citep{james,Cannon}. 
The central idea in the James--Lange theory is represented in the quote ``We don't laugh because we're happy, we're happy because we laugh'' by James. 
However, the Cannon--Bard theory contradicts it. 
The question of which of these theories is correct has long been controversial. 
Schachter, in contrast, advocated a two-factor theory and developed an emotional theory that included these two competitive theories \citep{Schachter}. 
Cognitive theory is also famous for incorporating cognitive activities in the form of judgments, evaluations, or thoughts \citep{Arnold1960,SLazarus1991,OCCbook}. 
These models give important implications for emotion; however, because they do not model the entire mechanism of emotions, they do not necessarily clarify what emotions are. 
Moreover, they are not computational models. 
In other words, there is also a problem that the models cannot be directly implemented on a computer. 

Neuroscience has revealed neural circuits, such as the Papez circuit \citep{papez1937} and Yakovlev circuit \citep{YAKOVLEV1948}, that are relevant to emotions. 
LeDoux discussed the function of the brain in emotions in detail based on the anatomical point of view. 
He proposed the dual pathway theory, which claims that there are two types of emotional processing paths: automatic and rapid processing by the limbic system, and complicated and higher cognitive processing from the neocortex to the amygdala \citep{LeDoux1986,LeDoux1989,ledoux1998emotional}. 
More recently, the quartet theory of emotions, which claims four important systems for emotions in the brain, was proposed \citep{KOELSCH20151}. 
These are the brainstem-centered, diencephalon-centered, hippocampus-centered, and orbitofrontal-centered systems. 
As a matter of course, the authors argue that the limbic/paralimbic structure, i.e., the basal ganglia, amygdala, insular cortex, and cingulate cortex, are also of importance for affective processes. 
As shown in computational neuroscience studies, the cortical-basal ganglia loop can be considered as a reinforcement learning module. 
In particular, the striatum plays a very important role in sensorimotor, attentional, and emotional processes. 
These neuroscientific findings are not only important for concretely considering emotion models, but also have direct implications on computational models. 

From the viewpoint of human--robot interaction, emotion is one of the most important factors for partner robots. 
The intuition that the difference between humans and AI (robots) is in emotion implies that, in other words, the realization of the emotion model may be the key to realize robots and/or AI with high affinity for human beings. 
Picard proposed the idea of affective computing, in which the emotion recognition in humans has been studied extensively, mainly by examining facial expressions \citep{AffectiveComputing}. 
The success of deep learning in recent years has accelerated this line of research.  
Of course, the classification of a person's inner state based on facial expressions is very useful for the robot to communicate with us, because it can select its response according to the recognized result. 
However, it is fair to say that the recognition of facial expressions is different from a ``true understanding'' of the emotional states of others, even though a highly accurate facial expression recognition method is available thanks to deep learning technologies. 
Robots need to understand, sympathize, and act according to their partners' complex emotional states in order to become accepted members of human society. 
Toward this goal, many efforts on designing emotional expressions for social robots have been made \citep{DesigningSociable}. 
However, almost all these emotions have been designed manually. 
High-level complex social emotions for robots are difficult to preprogram manually. 
In fact, conventional studies have only been able to accomplish simple basic emotions such as happiness and sadness \citep{RoboticEmotional,Verbalconversation}. 
An emotion model for robots based on the difference equation was proposed \citep{miwa01}; however, the system was too simple to generate complex higher-level emotions. 
%
The basic idea underlying this study is that the problem of emotions should be formulated as ``understanding by a generative process of emotions'' rather than ``classification.'' 
If we abandon the manual design of emotions, emotional differentiation \citep{Bridges,lewis2000self} must be the right path to follow in order to achieve this ultimate goal. 
This idea shares the same goal as the affective developmental robotics proposed by Asada \citep{ASADA15}. 

Accordingly, we propose a computational model of emotions, which is based on certain neurological and psychological findings in the literature. 
The purpose of this paper is first to present a general meta-level framework for the mechanism behind emotions. 
The literature on emotions in the past as discussed above motivates us to propose a three-layer model to cover emotions. 
The first layer corresponds to the appraisal module, which is responsible for quick evaluation of the external world and internal body. 
Interoception in particular, which is sensitivity to stimuli originating inside the body, is a very important factor. 
The second layer has an emotional memory to adjust the innate appraisal module in the first layer to the surrounding environment, which the agent is facing. 
The third layer includes reinforcement learning and sequence learning modules that correspond to the cortical-basal ganglia loop. 
This is because the important aspect of emotions is their role in decision-making \citep{moerland2017emotion}. 
The dual path theory by LeDoux is one of the important theories of emotions, which forms the basis of our proposed three-layer model. 
Moreover, this three-layer model roughly matches the recent neurobiological emotion model \citep{KOELSCH20151}. 

We also attempt to implement the three-layer model of emotion using deep neural networks. 
Our proposed implementation relies on a combination of the recurrent attention model (RAM) \citep{RAM2014} for the first layer, as well as the convolutional long short-term memory (LSTM) \citep{xingjian2015convolutional} and the reinforcement learning module using the deep deterministic policy gradient (DDPG) \citep{DDPG} for the third layer. 
The second layer is realized based on a mechanism of nonlinear smoothing, which makes the whole emotion system adaptable to the surrounding environment. 
%
%
Then, the implemented computational model of emotion is tested by employing certain tasks simulating mother-–infant interaction to evaluate the plausibility of the model. 
Some promising results are obtained in the experiment. 
For example, we found that the policy network represents emotional states and exhibits emotion differentiation in the proposed three-layer model. 
We believe this constructive approach toward emotions may yield a clue to the elucidation of human emotions. 
Moreover, the generative model of emotion is also important for achieving empathic communication between humans and robots. 

The contributions of this study are threefold. 
First, this study investigates a meta-level model of emotion as a whole. 
Second, an implementation of the emotion model using deep leaning modules is provided. 
Third, we design a simulation task mimicking mother--infant interaction in order to evaluate the model, which reveals that the proposed model is indeed able to show emotion differentiation. 
It should be noted that our previous studies showed some preliminary ideas and examinations of the proposed emotion model \citep{HRI2017,hieida2018decision,hieida2018ROMAN}. 
Although the basic idea of this work is shared with the previous studies, this paper provides a detailed explanation of the model and full implementation as an entire emotion network, which were not given in the previous works. 
Moreover, we repeated the entire experiments and the results presented in this paper are completely new. 

The remainder of this paper is organized as follows. 
In the next section, the literature on emotions is discussed and then a model of emotions is proposed. 
Section 3 provides an implementation of the proposed emotion model using deep neural networks. 
Each module of the network is explained in detail. 
The experiments are presented in Section 4, which indicate plausibility of the proposed deep emotion model. 
Finally, this paper is summarized in Section 5.

%% file: model.tex
\section{Model of Emotion}
Here, we present an overview of the basic idea of our proposed emotion model. 
Some important findings for our proposal are reviewed first, followed by the proposed model of emotions. 
\subsection{Emotions in literature}
What is emotion? 
In order to propose a model of emotion, we should start from this important question. 
Moreover, we need to clarify the definition of emotion. 
To the best of our knowledge, there is no universal consensus on the definition of emotion; however, recent research reveals the importance of the body in emotion. 
This is what William James claimed long ago, also called the peripheral origin theory of emotion \citep{james}. 
Recent studies in cognitive neuroscience have revealed that interoception, which is a perception of the internal bodily state, is a key for the subjective experience of emotion \citep{Terasawa2013humanbrainmapping}. 
In the quartet theory of emotions, the brainstem-centered system corresponds to this type of emotion system \citep{KOELSCH20151}. 
The brainstem is the oldest brain structure and the reticular formation plays an important role in this system. 
Another important aspect of embodiment in emotion is Damasio's somatic marker hypothesis, which hypothesized that emotions evaluate external stimuli efficiently through our own body \citep{Damashio}. 
This motivates us to consider both internal and external appraisals simultaneously. 

In any case, the physical body is an origin of emotions and is indispensable. 
In this paper, we consider the body and interoception as one subsystem, i.e., the first layer in the proposed model. 
In fact, despite the differentiating property of emotions, some basic emotions such as anger, joy, disgust, fear, sorrow, and surprise exist regardless of culture \citep{ekman}. 
This is possible because we as human beings share similar physical bodies and environments. 
This result supports the fact that emotions are based on our physical bodily states. 
The idea of active inference is also related to this embodied system, e.g., visual attention is relevant to the appraisal of visual stimuli \citep{Friston2010,Seth20160007}.

Another important aspect of emotion is related to decision-making \citep{Ledoux} and inference on causal attribution. 
For example, a misattribution of arousal, which is also known generally as the suspension-bridge effect, has been found to happen to someone who experiences the effects of fear of physical danger while meeting someone, and who mistakenly believes that the other person is the cause of their physical responses \citep{Dutton}. 
This higher-level cognitive process seems to be deeply related to the orbitofrontal-centered system. 
The reinforcement learning module, which originates from the cortical-basal ganglia loop, is also related. 
The relationship between active inference and reinforcement learning has been discussed \citep{Friston2009PLOS}, which implies this system is also related to the active inference. 
In this paper, we consider the decision-making as another subsystem, which we will call the third-layer in the proposed model. 

The discussion so far implies that one's emotional system is divided into two systems: 1) the hardwired innate system and 2) the learning system, which is related to the decision-making. 
Now, we define emotions (emotional states) and feelings (emotional feelings). 
Emotions are defined as a set of physical reactions, state changes of visceral and skeletal muscles, and changes in internal conditions. 
These changes are evoked by the above systems 1) and 2). 
In contrast, feelings are defined as perceptions of the emotional states. 
These definitions are based on Damasio's definition \citep{damasio2003looking}. 
It should be noted that the term ``emotion'' corresponds to ``affect'' in the area of psychology: this paper uses the term ``emotion,'' which is generally acceptable. 
\begin{figure}[h!]
\begin{center}
\includegraphics[width=160mm]{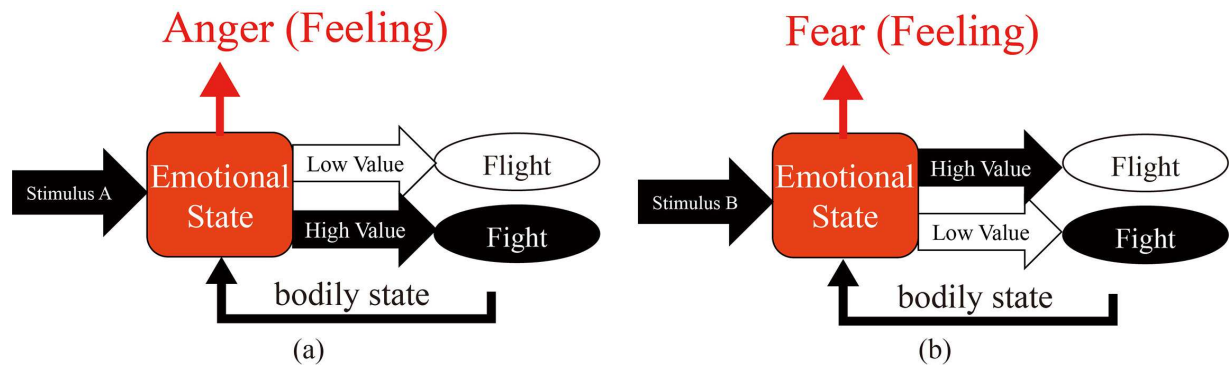}
\caption{Illustration of ``anger'' and ``fear,'' which highlights the difference: (a) emotional feeling of anger, and (b) emotional feeling of fear.}
\label{fig:exam}
\end{center}
\end{figure}

In order to clarify the definition of emotions/feelings used in this paper, Fig. \ref{fig:exam} illustrates concrete examples. 
In the figure, there are a stimulus A and a bodily state that evoke the ``Fight'' action, whereas a stimulus B and a bodily state activate the ``Flight'' action. 
In this case, the emotional state that stimulus A and the bodily sate cause is labeled as ``anger,'' and the emotional state caused by the stimulus B and the bodily sate is labeled as ``fear.'' 
This definition directly connects emotions to the somatic marker hypothesis, which means that the emotion should be generated by considering internal appraisal, external appraisal, and decision-making mechanisms. 

Regarding the learning system, a memory-based system is an important candidate as a building block of the emotion model. 
In the quartet theory, the hippocampus-centered system corresponds to the memory-based system, in which the hippocampus and amygdala are mainly involved \citep{KOELSCH20151}. 
The activity of the amygdala in emotion is particularly important and has been studied for a long time. 
Yakovlev's circuit is one of the well known limbic systems and the amygdala is involved in the circuit \citep{YAKOVLEV1948}. 
Papez's circuit is another well known limbic circuit, which includes the hippocampus \citep{papez1937}. 
Although these are independent as circuits, they have mutual interaction and are closely related each other through the cortex, basal ganglia, and diencephalon \citep{mendoza2007clinical}. 
In this paper, we consider the memory-based system as another subsystem, which we call the second layer in the proposed model. 
This subsystem gives flexibility to the innate-appraisal system, i.e., first layer, in order to adapt the whole system to the environment. 

Eventually, emotions cannot be viewed locally, and need to be thought of as a network. 
Therefore, the abovementioned subsystems should be connected as a network to generate emotions. 
Furthermore, the important aspect of the model is its ability to explain various phenomena known in the art. 
Among others, emotion differentiation is an important phenomenon, because it is a key to implementing emotions for robots, as mentioned earlier. 
Bridges claimed that excitement, which is the origin of emotion, can be divided into several emotional categories based on observations of infants \citep{Bridges}. 
More recently, it has been reported that emotions such as pleasure, interest, surprise, sadness, anger, and fear were recognized one year after birth; pride, shame, guilty feelings, etc. emerged from approximately two and a half years; and not all but great majority of emotions appear by the age of three \citep{lewis1995}. 
Our proposed model of emotion is discussed in the next subsection. 
Then, the model is implemented using deep neural networks and tested to determine whether it develops the emotional categories.  
%

\subsection{Proposed model of emotion}
\begin{figure}[h]
\begin{center}
\includegraphics[width=150mm]{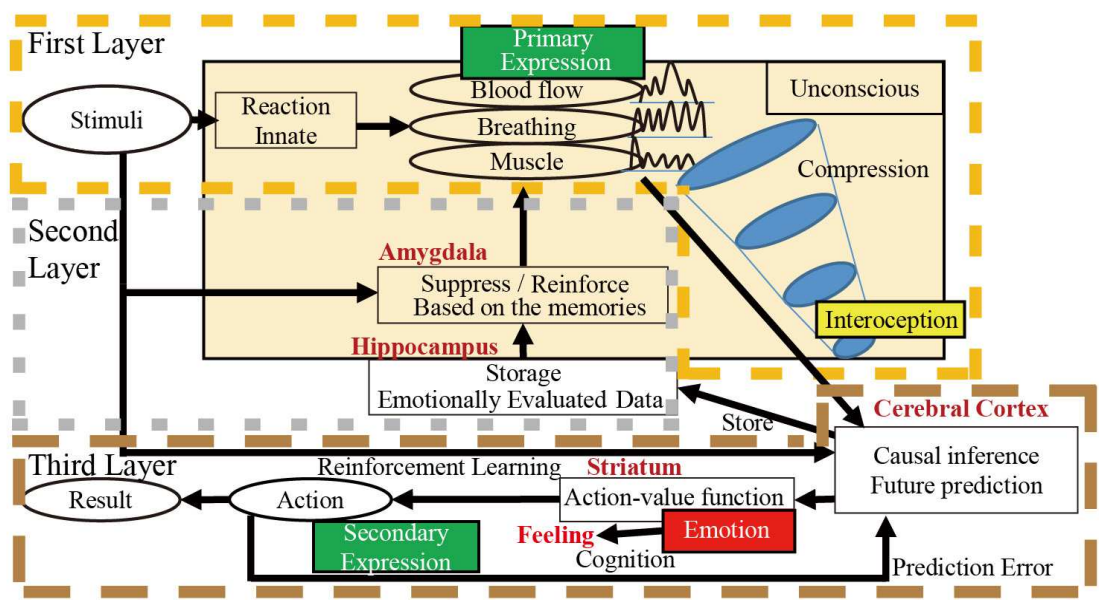}
\caption{Schematic diagram of our proposed three-layer model of emotion.}
\label{fig:model}
\end{center}
\end{figure}

The proposed emotion model is illustrated in Fig. \ref {fig:model}. 
The emotion model is divided into three layers: the first layer that reacts bodily to stimuli very fast, the second layer that accesses memories such that stimuli can be evaluated through experiences, and the third layer that makes future predictions and actions. 
These are derived from the abovementioned implications. 

The first layer reacts to stimuli very quickly using the body, which is called external appraisal. 
Moreover, this part reflects the situation of the body itself, i.e., internal appraisal, regardless of external perception. 
This layer is the reason why emotions depend on the physical body. 
Because the reactions are preprogrammed innately, they usually contain errors, which cause overreactions to stimuli. 
To alleviate this problem, the second layer accesses memories such that stimuli can be evaluated through experiences. 
This second layer makes it possible to suppress unnecessary reactions and, at the same time, react quickly to important problems. 
Of course, this is a trade-off between processing cost and accuracy of response to stimuli. 
Hence, the output of the first layer, which is modulated by the second layer to be precise, can be considered as the perception of dimensionally reduced evaluated results of the 
external and internal worlds, i.e., internal representation.  
Therefore, the perception of the output of the first layer can be regarded as interoception. 

In the third layer, the output of the first layer is used together with the input stimuli for causal inference and prediction, as shown in Fig. \ref {fig:model}. 
Subsequently to the prediction, decision-making is carried out using the input stimuli and the results of the prediction. 
The most important part of the third layer is reinforcement learning, which is responsible for the learning of optimal decision-making. 
One of the most important aspects of the reinforcement learning is the definition of a reward. 
In the model of emotion, the idea of ``homeostasis,'' which is a regulatory mechanism of the agent's internal state, should be adopted. 
This is based on the drive reduction theory, which is the basic theory of motivation \citep{mayer2010}.
It interesting that homeostasis is closely related to the diencephalon, which is one of the emotion systems in the quartet theory of emotions \citep{KOELSCH20151}. 
Hence, a reward is provided when the output of the first layer, i.e., interoception, remains constant. 
This constant is not a completely constant value. 
It takes the average value of emotional state over a time window with a certain length.
In other words, homeostasis is set not to keep the emotional state completely constant, but to discourage rapid changes.
In our model, the average value that gradually changes in time is defined as ``mood.'' 
Thereafter, the neural patterns of the policy in the striatum, i.e., emotional states, are consciously recognized as emotional feelings. 
After the decision-making process, the prediction error is calculated followed by updating of the model in the third layer. 
Experiences are stored in the hippocampus as episodic memories, with emotional evaluation in the second layer. 
It is worth noting that the learning process exists only in the second and third layers. 
\begin{figure}[h]
\begin{center}
\includegraphics[width=150mm]{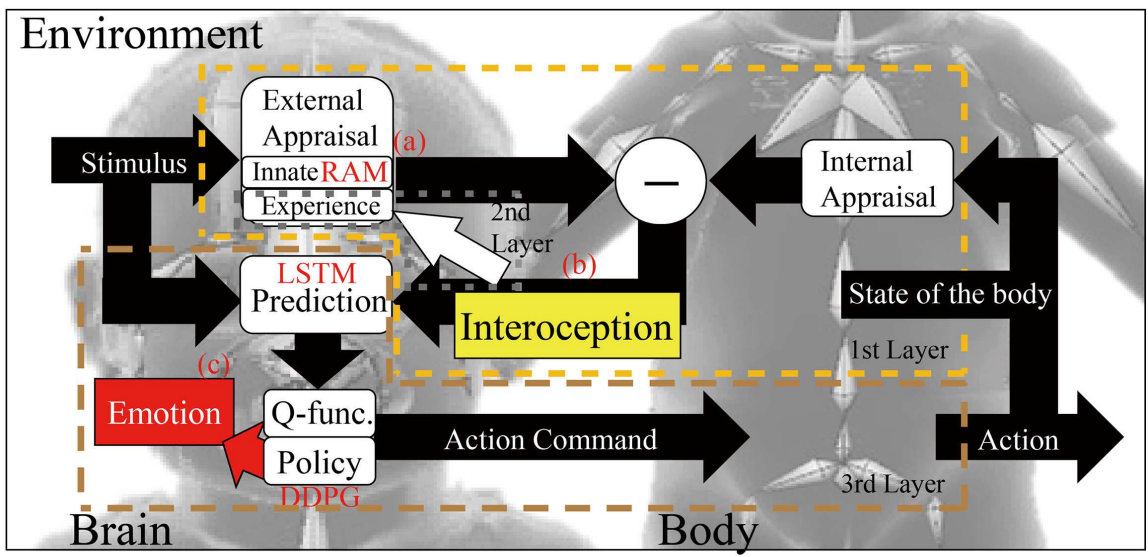}
\caption{Our proposed emotion model for implementation, which is a redrawn version of Fig. \ref{fig:model}.}
\label{fig:model-impl}
\end{center}
\end{figure}

Figure \ref{fig:model-impl} is a redrawn version of the proposed model in Fig. \ref{fig:model}, in order to make it comprehensive for implementation. 
This figure directly claims some important points of our proposal. 
First, the internal and external appraisals, i.e., embodiment, are the sources of emotions. 
It is fair to say that without the physical body, there should be no emotions. 
Second, prediction is indispensable in the proposed model. 
Third, another key point in our model is the decision-making part, which is relevant to the somatic marker hypothesis. 
These viewpoints remind us to note the close relationship between our proposed emotion model and the embodied predictive interoception coding (EPIC) model, which was proposed recently \citep{EPIC2015}. 
The idea of the EPIC model is based on predictive coding and active inference \citep{Friston2007}. 
Although we developed our proposed model independently of the EPIC model, some important ideas are shared between the two models. 
The main difference between the EPIC model and the model proposed in this study is that we propose the actual implementation of the proposed model by combining several deep learning modules, which are described in the next section. 
On the contrary, the EPIC model is a conceptual model and sticks firmly to the predictive coding. 

Another important aspect for the model is the design of artificial emotional systems, which Ca{\~n}amero contends \citep{canamero2005emotion}. 
She claimed that emotions must be grounded in an internal value system that is meaningful for the robot's physical and social niche. 
The model should establish a link between emotions, motivation, behavior, perception, and various aspects of ``cognition,'' and the link must be rooted in the body of the agent. 
As already discussed, our proposed emotion model has the potential to fulfill these requirements. 

%

%% file: implement.tex
\section{Implementation}
This section proposes an implementation of the emotion model described in the previous section. 
The proposed implementation consists of a combination of deep neural networks, such as RAM, LSTM, and DDPG, except for the second layer. 
The second layer is realized by a simple smoothing mechanism to make the learning system tractable. 
In the following, we will look at the implementation of each module in turn. 

\subsection{Appraisal module (1st layer)}
\label{1stlayer}
As we discussed earlier, the first layer is responsible for generating interoception based on both internal and external appraisals. 
The problem here is the generation of a physical response to stimuli. 
The model is required to generate a ``human-like'' response in order to replicate human emotions. 
In this study, we attempt to generate suitable affect values, i.e., a pair of valence and arousal values, from input visual stimuli using a neural network instead of generating physical body reactions \citep{HRI2017}. 
To replicate human-like innate reactions, we utilize several databases such as the international affective picture system (IAPS) database \citep{IAPS,lang1999international}, the open affective standardized image set (OASIS) \citep{kurdi2017introducing}, the Nencki affective picture system (NAPS) \citep{marchewka2014nencki}, and the Geneva affective picture database (GAPED) \citep{dan2011geneva}, to train the network in order to generate two-dimensional valence and arousal values for a given visual stimulus. 
Therefore this is a regression problem. 

\begin{figure}[h]
\begin{center}
\includegraphics[width=170mm]{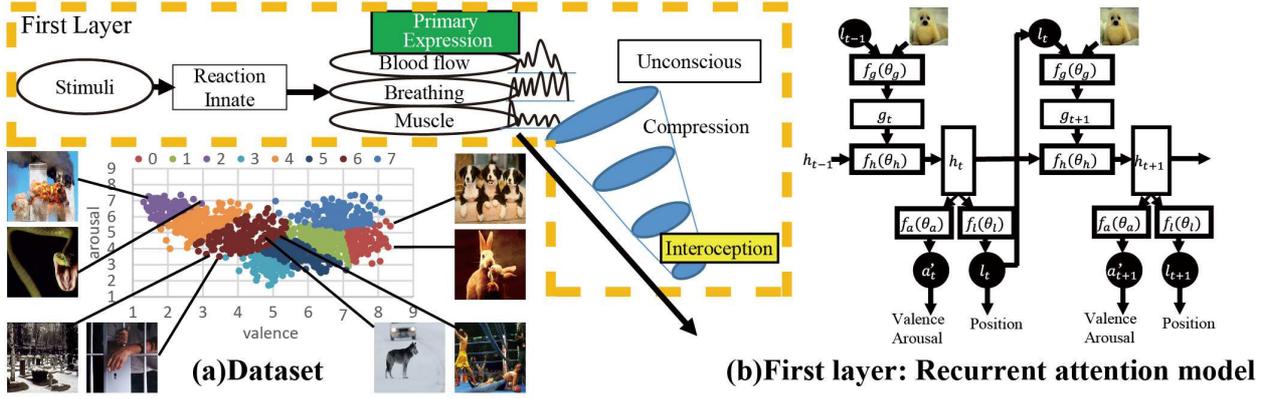}
\caption{Overview of the first layer implemented by the recurrent attention model: (a) block diagram of the first layer and image examples from IAPS \citep{IAPS}, and (b) network architecture of the RAM \citep{HRI2017}. }
\label{fig:model-ram}
\end{center}
\end{figure}

To take the active interoceptive inference into consideration, we propose the use of the RAM \citep{RAM2014}. 
This is because visual attention is a very important factor for estimating arousal and valence values, and the RAM makes it possible to learn the visual attention and affect values simultaneously, as shown in Fig. \ref {fig:model-ram}. 
Please refer to the appendix for details on the RAM. 
It should be noted that the RAM improved the performance of regression compared with the convolutional neural network (CNN), which is directly trained using pairs of images and ground truth. 
%
%
%

\subsection{Emotional memory module (2nd layer)}
\begin{figure}[h]
\begin{center}
\includegraphics[width=170mm]{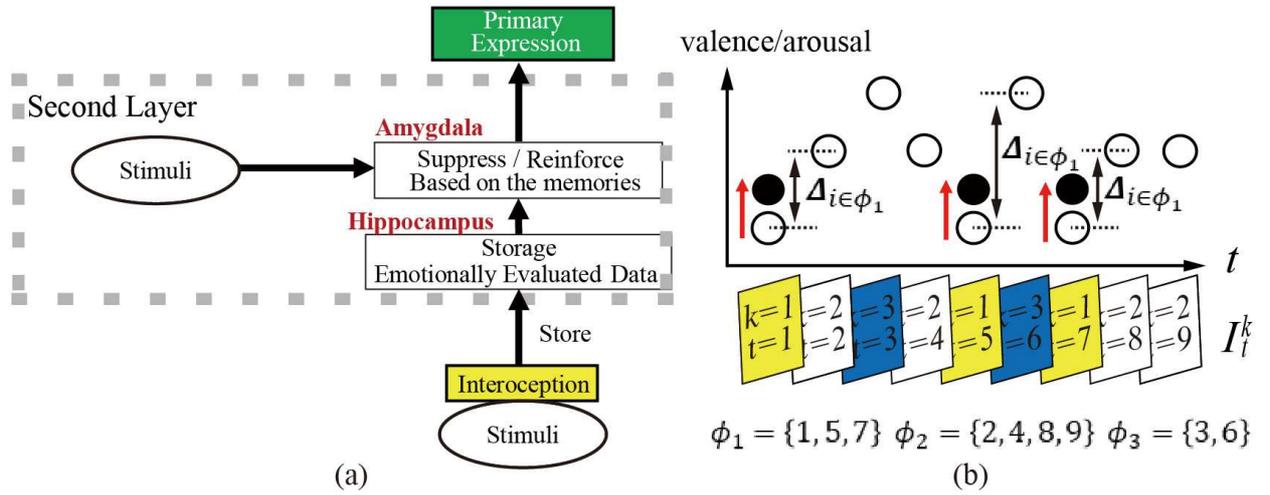}
\caption{Second layer implemented by the smoothing: (a) block diagram of the second layer, and (b) schematic example of the smoothing process. Because $\bm{\Delta}_{i \in \phi_1}$ are always positive, the compensation term $\bm{L}(1)$ is positive in this example. This means that the input stimuli belonging to the category $k=1$ reinforce high affect values. 
The white circles for $t=1, 5, 7$ are moved to the black circles by the compensation term $\bm{L}(1)$.}
\label{fig:model-smm}
\end{center}
\end{figure}
The second layer shown in Fig. \ref {fig:model-smm} (a) can be regarded as an adaptation using data in the actual environment for the innate and fixed system, i.e., the first layer. 
The memory-based learning increases the accuracy of the prediction by using the past accumulated information experienced by the agent. 
Here, we formulate this as a problem of calculating the expected value $E\left [x(t)|\bm{z}_0^T \right ]$ ($0 \leq t \leq T $), where $x(t)$ and $\bm{z}_0^T$ represent the target value to be estimated and the stored data, respectively. 
This is a type of smoothing problem and the second layer is realized by a simple nonlinear smoothing technique, as shown in Fig. \ref {fig:model-smm} (b). 

More specifically, a time series including affect values and stimuli during a certain period of time is stored in the memory. 
The idea here is that the output of the RAM is modified by the compensation term $\bm{L}(\cdot)$ in the second layer as follows: 
\begin{eqnarray}
\bm{a}'(t) & =& RAM(I_{t}^{k}) + \bm{L}(k),\\
\bm{a}(t) &= &\bm{a}'(t) + IA(t), 
\label{eq:a(t)}
\end{eqnarray}
where $\bm{a}'(t)$ is an external appraisal at time $t$, 
$RAM(I_{t}^{k})$ represents the output of the first layer for the input image $I^k_t$ at time $t$, 
$k$ indicates the category of the input image, 
$IA(t)$ is an internal appraisal at time $t$, which will be described later,
and $\bm{L}(k)$ represents the output of the second layer (compensation term) for the image $I_t^k$. 
$\bm{L}(k)$, which modifies the first layer output $RAM(\cdot)$, is updated using the stored data as follows: 
\begin{equation}
\bm{L}(k) \leftarrow \bm{L}(k) + \gamma \frac{1}{|\phi_k|} \sum_{i \in \phi_k}\left\{ \bm{a}(i+1)-\bm{a}(i)\right\} = \bm{L}(k)+\gamma \frac{1}{|\phi_k|} \sum_{i \in \phi_k} \bm{\Delta}_{i \in \phi_k}, 
\end{equation}
where $\gamma$ is the learning rate and is set to 0.1 in the later experiment. 
$\phi_k$ is a collection of time indices $t$ with the same image category $k$ and $|\phi_k|$ represents the number of images belonging to $\phi_k$. 

As shown in Fig. \ref {fig:model-smm} (b), the smoothing process can capture temporal information. 
For example, when the next affect values for a particular image category $k=1$ increase frequently, the term $\bm{L}(1)$ compensates the affect value of the corresponding image input in the upward direction. 
However, the next affect values for $k=2$ vary and the sum of $\bm{\Delta}_{i \in \phi_1}$ cancels out. 
With this smoothing process, we can expect the effect of lowering the load on the body, and it improves the prediction performance of the next layer. 

Long-term potentiation (LTP) is a well known mechanism for connecting memory and learning. 
The hippocampus and amygdala are closely related to the LTP mechanism. 
The smoothing mechanism in this layer is assumed to mimic LTP in the functional level. 
Moreover, the amygdala is involved in the classical conditioning based on LTP. 
Thus, the process of this layer may replicate the classical conditioning. 
Fig. \ref {fig:model-smm} (b) also explains this mechanism in a simple way. 
The second layer learns that the image category $k=1$ is the trigger of high valence and arousal values. 
%

The output of RAM and second layer represents external appraisal. 
According to our definition, interoception is a combination of external appraisal, i.e., the output of RAM modulated by the second layer, and internal appraisal representing internal energy, as shown in Fig. \ref{fig:model-impl}. 
The internal energy increases or decreases according to the selected action. 
For example, moving the body forcefully consumes energy, consequently the internal energy decreases (the internal appraisal increases). 
Because the internal appraisal depends on the definition of the agent to be assumed, we explain certain details on the implemented internal appraisal module in the next subsection. 

\subsection{Internal appraisal}
In this study, the internal appraisal module is implemented in a rule-based manner. 
Essentially, the internal appraisal increases when the agent acts as the internal energy is decreased. 
When the agent shows sadness or closes his eyelids, the internal appraisal decreases. 
This is because we assume that showing sadness leads to getting milk, and closing his eyelids corresponds to sleeping, which restores physical strength. 
The internal appraisal implies a physical strength bias in general, and the interoception is expressed by applying the physical strength bias to the external appraisal, as shown in Fig. \ref{fig:model-impl}. 
These assumptions are made because of a mother--infant interaction scenario in our later experiment. 
The agent has four facial parts to move according to external stimuli. 
Each facial part can be continuously controlled by the agent at the cost of corresponding power consumption. 
Thus, the agent has to learn (in the third layer) suitable facial expressions according to the external and internal worlds. 
More precisely, the internal appraisal $IA(t)$ can be rewritten as the following formula: 
\begin{eqnarray}
IA(t) &=& \sum_{n=0}^{3} \left( 1-exp \left\{ -\frac{A_{n}(t)}{\tau} \right\} \right),  \label{eq:IA} \\
A_{n}(t) &=& 
	\left\{
    		\begin{array}{l}
      			|A_{n}(t-1)-d|   ~:for~closing~eyelids~actions~or~showing~sadness\\
			A_{n}(t-1)+a^c_{t-1}+\eta  ~: otherwise
    		\end{array},
  	\right. \label{eq:A_{n}(t)}
\end{eqnarray}
where $a^c_{t-1}$ is an action cost at time $t-1$, and $n=0, \cdots, 3$ represents facial parts. 
$\eta$ represents constant physical fatigue and is set to $0.01$ in the later experiments. 
As four facial parts are assumed, $IA(t)$ has a value in the range from 0 to 4. 
Eq. (\ref{eq:IA}) denotes a basic curve of physical strength with a time constant $\tau$ ($\tau=50$ in the later experiments). 
Eq. (\ref{eq:A_{n}(t)}) represents change in the parameter of the basic curve. 
It is natural that the parameter $A_{n}(t)$ increases as the action is taken by the agent. 
When the agent closes his eyelids, the parameter is set such that the physical strength recovers. 
Additionally, even when the agent expresses a sad expression, the parameter is set for restoring physical strength. 
As we mentioned earlier, these settings are based on the assumption of mother-–infant interaction. 
In this study, $d$ is set to $50$ for closing eyelids and 75 for showing sadness. 
If these two values are the same, the two types of actions become meaningless. Therefore, they are set to different values, such that each action becomes meaningful in the reinforcement learning module. 
It should be noted that it is possible to design other rule-based internal appraisal modules according to the physical body of the agent and the scenario of the world in which the agent exists. 
\subsection{Decision-making module (3rd layer)}
\begin{figure}[htb]
\begin{center}
\includegraphics[width=170mm]{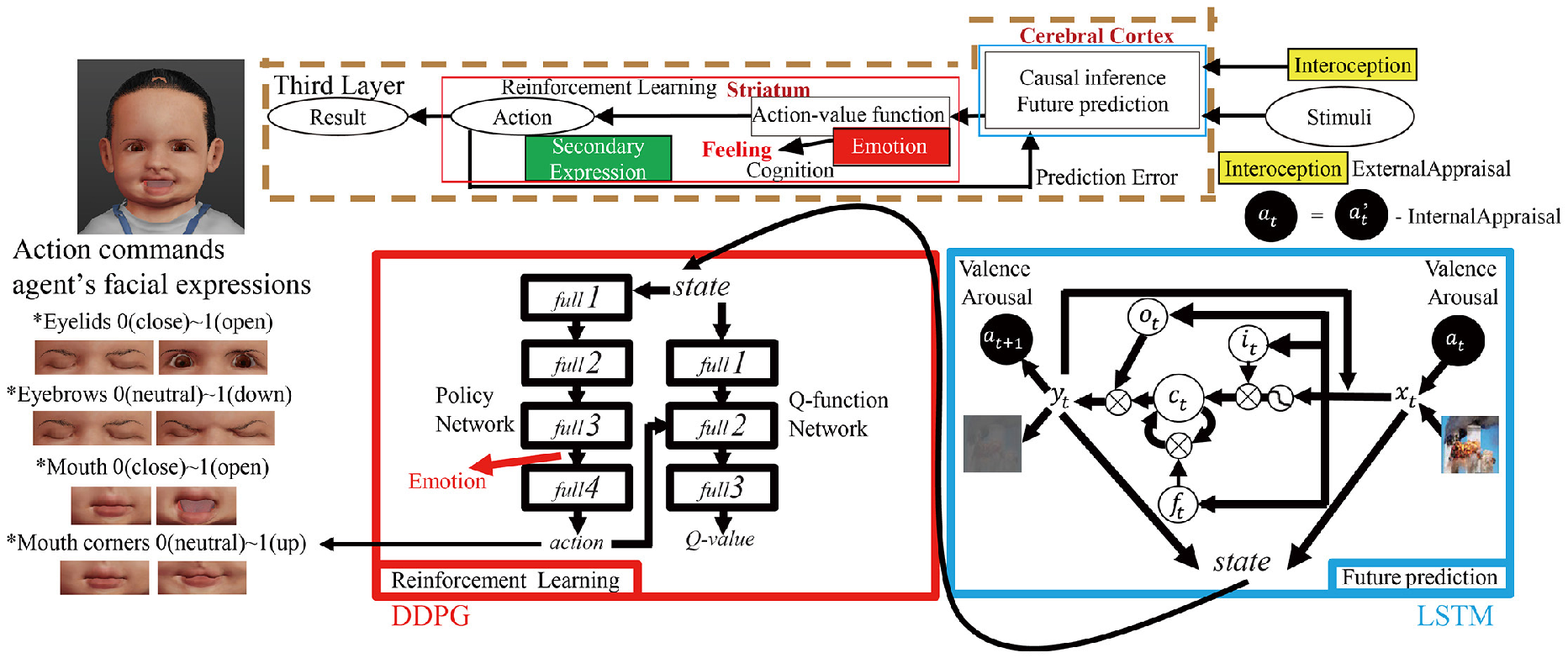}
\caption{Overview of the third layer implemented by convolutional LSTM  and DDPG: schematic of the third layer, network architecture, and infant agent used in the later experiments.}
\label{fig:model-lstm-ddpg}
\end{center}
\end{figure}
As shown in Fig. \ref {fig:model-lstm-ddpg}, the decision-making module is implemented using convolutional LSTM and DDPG. 
In a previous study, we have implemented using LSTM--DQN \citep{hieida2018decision}.  
The LSTM--DQN has a drawback that continuous actions cannot be dealt with. 
That is why the convolutional LSTM--DDPG is employed in this study. 
Please refer to the appendix for details on the convolutional LSTM and DDPG. 
To train the network (reinforcement learning), combinations of an input image such as Fig. \ref {fig:model-ram}(a), and the result of subtraction between an output of the RAM and an internal appraisal, i.e., interoception, are used. 
Another important part of reinforcement learning in general is the actions. 
This means that the implementation of the proposed emotion model requires actions, because the reinforcement learning is employed. 
Here, we discuss the actions used in this study. 
To consider the actions in the reinforcement learning, we need to assume the robot/agent to be used, because the actions to take vary depending on the body of the robot/agent. 
Without loss of generality, we assume the agent that is used in our later experiment in this study. 
The agent has action commands of its own facial expressions for given visual stimuli and interoception in the first layer, i.e., valence and arousal values. 
The convolutional LSTM is responsible for predicting an image and interoception values at the next time-step from the input image and current interoception values. 
The DDPG module generates an action command by taking the input image, interoception values, and predicted results by the convolutional LSTM, as an input.  
Figure \ref{fig:model-lstm-ddpg} illustrates the overall processing of the decision-making module (third layer). 

As discussed in the meta-level model, the idea of homeostasis is used for calculating the reward as follows: 
\begin{eqnarray}
  R(t)& =& C - {\left\| \bm{m}(t) - \bm{a}(t) \right\|}^2_2 , \\
  \bm{m}(t)& =& \frac{1}{2}\left ( \bar{\bm{a}} + \frac{1}{N}\sum_{i=1}^{N} \bm{a}(t-i) \right),  \label{eq:mt}
\end{eqnarray}
where $R(t)$ and $\bm{a}(t)$ represent, respectively, the reward value and the vector consisting of valence and arousal values, i.e., interoception values, at time $t$. 
$\bm{m}(t)$ represents the mood of the agent at that moment and is calculated as a mean vector of $\bar{\bm{a}}$ and the average of the past $N$ frames. 
$\bar{\bm{a}}$, $N$, and $C$ represent a vector consisting of intermediate values between maximum and minimum interoception values, number of averaging frames, and a constant value, which translates the differential value to a reward value. 
Eq. (\ref{eq:mt}) is intended to represent a mood, which is less likely to be provoked by a particular stimulus and is determined by the average of the last $N$ interoception values. 

\subsection{Learning of the model}
\begin{figure}[h]
\begin{center}
\includegraphics[width=150mm]{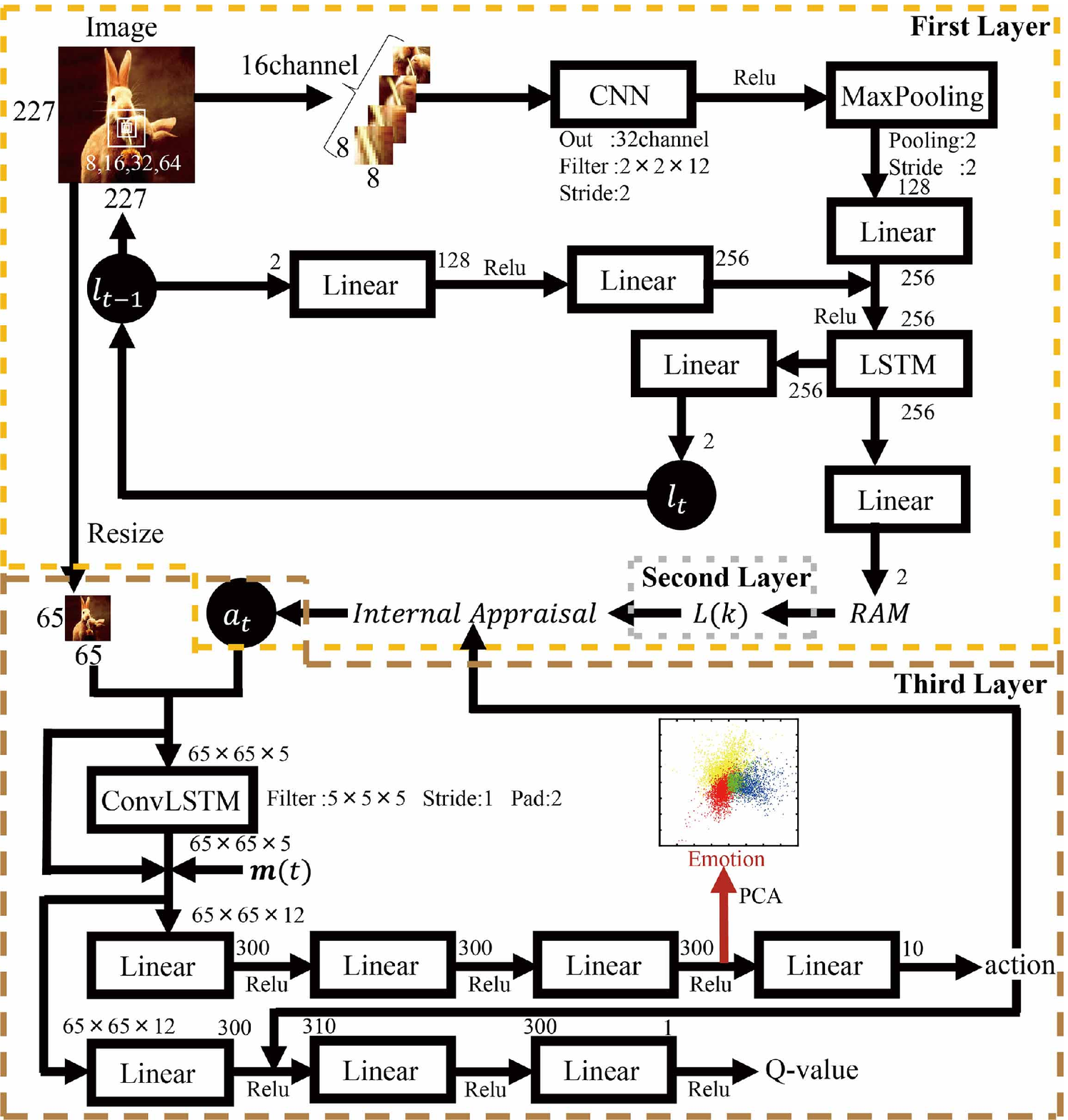}
\caption{Whole network architecture of the proposed deep emotion.}
\label{fig:appimpl}
\end{center}
\end{figure}

Because our proposed model consists of several learning modules, several patterns can be considered as the timing of these updates. 
This study takes a simple idea of updating the LSTM and the second layer at each timing based on DDPG update loop. 
The entire learning algorithm of the proposed model is shown in Algorithm \ref{alg2}. 
In the algorithm, we set two parameters empirically as $T_{LSTM}=100$ and $T_{L2}=1000$. 
Figure \ref{fig:appimpl} shows the whole network architecture of the proposed model. 
One can see the detailed parameters, such as number of input/output nodes, in the figure. 

\begin{algorithm}
\caption{Deep emotion learning algorithm}         
\label{alg2}                          
\begin{algorithmic}
\STATE Train the recurrent attention model $RAM(\cdot)$ (offline)
\STATE Initialize the mood of the agent ${\bm m}(0)$
\STATE Initialize the second layer $\bm{L}(k)$
\STATE Randomly initialize critic network $Q(s,a|\theta^{Q})$ and actor $\mu(s|\theta^{\mu})$ with weights $\theta^{Q}$ and $\theta^{\mu}$        
\STATE Initialize target network $Q'$ and $\mu '$ with weights $\theta^{Q'} \leftarrow \theta^{Q}, \theta^{\mu '} \leftarrow \theta^{\mu}$
\STATE Initialize replay buffer $B$
\STATE Initialize a random process $\mathcal{N}$ for action exploration
\STATE Receive an initial input image $I_{0}^{k}$
\STATE Calculate interoception ${\bm a}(0)$ using Eq.(\ref{eq:a(t)})
\STATE Predict next image $\bar{I}_{0}^{k}$ and interoception $\bar{{\bm a}}(0)$ by LSTM module
\STATE Set $s_1 =  \left \{ I_{0}^{k}, {\bm a}(0), \bar{I}_{0}^{k}, \bar{{\bm a}}(0) \right \}$
\FOR{$e$ = 1, $M$}
\STATE Select action $a_{e} = \mu(s_{e}|\theta^{\mu}) + \mathcal{N}_{t} $ according to the current policy and exploration noise
\STATE Execute action $a_{e}$ and observe reward $R(e)$
\STATE Receive an input image $I_{e}^{k}$
\STATE Calculate interoception ${\bm a}(e)$ using Eq.(\ref{eq:a(t)})
\STATE Predict next image $\bar{I}_{e}^{k}$ and interoception $\bar{{\bm a}}(e) $ by LSTM module
\STATE Set $s_{e+1} =  \left \{ I_{e}^{k}, {\bm a}(e), \bar{I}_{e}^{k}, \bar{{\bm a}}(e) \right \}$
\STATE Store transition ($s_{e}, a_{e}, R_{e}, s_{e+1}$) in $B$
\STATE Sample a random minibatch of $N_B$ transitions ($s_{i}, a_{i}, R_{i}, s_{i+1}$) from $B$
\STATE Set $y_{i}=R_{i}+\gamma Q' \left ( s_{i+1},\mu '(s_{i+1}|\theta^{\mu '})|\theta^{Q'} \right ) $
\STATE Update critic by minimizing the loss: $L=\frac{1}{N_B}\sum_{i} \left \{ y_{i}-Q(s_{i},a_{i}|\theta^{Q}) \right \}^{2}$
\STATE Update the actor policy using the sampled policy gradient:\\ $\nabla_{\theta^{\mu}}J \approx \frac{1}{N_B}\sum_{i}\nabla_{a}Q(s,a|\theta^{Q})|_{s=s_{i},a=\mu(s_{i})}\nabla_{\theta~{\mu}} \mu(s|\theta^{\mu})_{s_{i}}$
\STATE Update the target networks:\\ $\theta^{Q'} \leftarrow \zeta \theta^{Q}+(1-\zeta)\theta^{Q'}$\\ $\theta^{\mu'} \leftarrow \zeta \theta^{\mu}+(1-\zeta)\theta^{\mu'}$
\STATE Store the loss of LSTM
\STATE Store interoception value and image for the second layer and the mood
\IF{$e$ is divisible by $T_{LSTM}$}
\STATE Update LSTM module
\ENDIF
\IF{$e$ is divisible by $T_{L2}$}
\STATE Update the mood of the agent ${\bm m}(t)$
\STATE Update the second layer $\bm{L}(k)$'s
\ENDIF
\ENDFOR       
\end{algorithmic}
\end{algorithm}

%

%% file: experiment.tex
\section{Experiments}
We explain the experiment in this section.
The experiment is roughly divided into three parts. 
In the first experiment, we verify the performance of the RAM (first layer). 
Because the first layer is assumed to return innate responses to stimuli, it is qualitatively evaluated through our subjective sense and children's tendencies. 

In the second experiment, we combined the RAM (first layer) and convolutional LSTM--DDPG (third layer), and observe the agent's behavior and internal representation of the emotional state. 
Because the second layer is responsible for the adaptation of the system to the environment, we focused on the implementation of first and third layers in this experiment.
In the third experiment, we combined the first, second, and third layers, implemented the whole emotion model, and verified its behavior.
Then, by comparing with the second experiment, the significance of the second layer is examined. 
\subsection{Experiments on RAM (1st layer)}
In order to test the performance of the RAM, we conducted the following experiment. 
\subsubsection{Experimental setup}
As explained in \ref{1stlayer}, the RAM was trained using a set of images from IAPS, OASIS, NAPS, and GAPED. 
We used in total 24,270 images (4,045 original images $\times$ 6 types of deformation such as rotations, flipping, and affine transformations) for training, and 100 images for testing (randomly selected from IAPS). 
After the training, the RAM was evaluated using the evaluation data.
To qualitatively examine the property of the model, we also input single-color images to the RAM and observed the results. 
This evaluation is expected to provide an insight on the color preference of the trained network. 
Moreover, we also input face images with a certain facial expression to the RAM. 
The Japanese female facial expression (JAFFE) database \citep{Dailey2010EvidenceAA} was used in this experiment. 
The JAFFE database contains 213 images of seven facial expressions (pleasure, sad, angry, fearful, surprised, disgusted, and neutral). 
We visualized outputs from the RAM, i.e., valence and arousal values, for these input images.  
\subsubsection{Results}
\begin{figure}[t]
\begin{center}
\includegraphics[width=150mm]{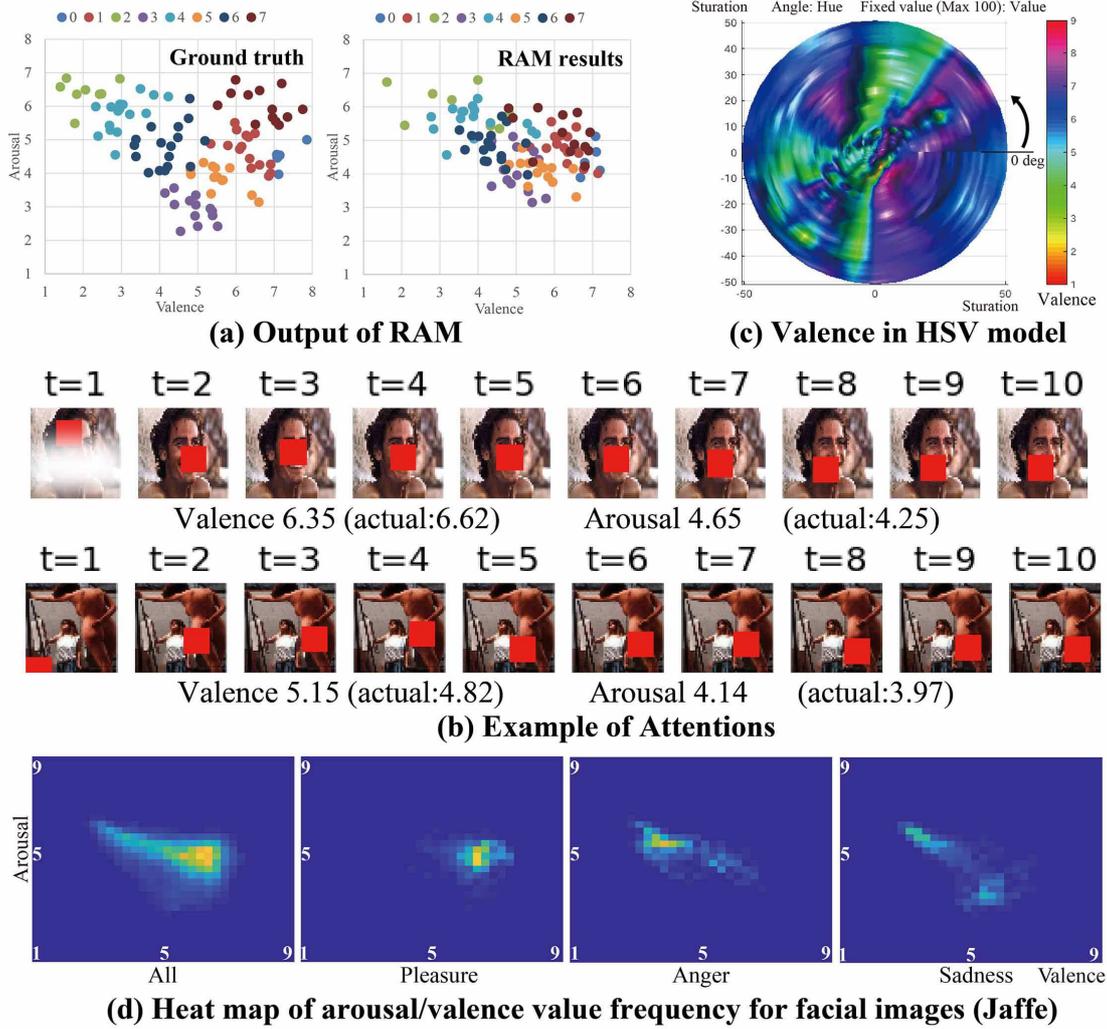}
\caption{Results of the first layer: (a) comparison between the output of the RAM and ground truth, (b) examples of the locations paid attention by the RAM (the red rectangle in each image represents the location of attention and a part of the facial image was blurred to make it impossible to identify individuals), (c) visualization of the RAM's output for input single-color images, and (d) heat map of arousal/valence frequency for facial images. }
\label{fig:ram-result}
\end{center}
\end{figure}

Fig. \ref{fig:ram-result} (a) represents the ground truth, i.e., values from the IAPS database, and the results output by the RAM. 
The mean absolute errors for 100 test images are 0.48 for arousal and 0.46 for valence. 
These errors are sufficiently small as compared with the variation of human evaluation \citep{IAPS,lang1999international}. 
Fig. \ref{fig:ram-result} (b) represents the results of visual attention for two different test images. 
One can see that the system successfully paid attention to visually important locations and estimated reasonable arousal and valence values in both cases. 

Fig. \ref{fig:ram-result} (c) shows the results of inputting single-color images; high values are observed around 45 degrees of hue, which corresponds to the color yellow.  
Additionally, high values are seen in the center, which corresponds to the color white. 
On the other hand, low values are observed around 270 and 100 degrees of hue, which correspond to purple and green, respectively. 
According to Yamawaki, for infants six months of age, warm colors, such as yellow, white, and pink, have high preference and cold colors, such as blue, green, and violet, have low preference \citep{color2010}. 
This result implies that the output of the RAM shows reasonable reactions compared with an infant. 

In the result of the facial image input, the output, i.e., valance and arousal values, tends to coincide with the category of the facial expression. 
The results are given in Fig. \ref{fig:ram-result} (d). 
For instance, when the facial images with pleasure expression are input to the RAM, the output of the RAM tends to have a high valence value. 
However, anger facial expressions tend to draw low valence and high arousal values. 
These results indicate that a response called emotional contagion \citep{Hatfield93,Barsade02} is observed in the trained RAM network.

\subsection{Experiments on decision-making module (3rd layer)}
\label{sec:1-3}
\subsubsection{Experimental setup}
This experiment intends to show the performance of the decision-making mechanism in the proposed emotion model. 
Therefore, we connected the RAM and the third layer, which is implemented by the convolutional LSTM--DDPG. 
The second layer is not included in this experiment, because the whole emotion model is used in the next experiment and the results are compared to examine the importance of the second layer. 
The virtual agent (we use a free software package called ``MakeHuman'' for the modeling of 3-dimensional agent\\ \url{http://www.makehumancommunity.org/wiki/MakeHuman_resources}), which can change its facial expressions by moving eyelids, eyebrows, mouth, and corners of the mouth, is used as the body and the RAM and the convolutional LSTM--DDPG are implemented in the virtual agent. 
Then, we designed a ``facial expression'' task based on the mother--infant interaction scenario. 
In this task, the interaction partner, which is also a computer agent (mother agent), recognizes the agent's facial expressions as one of four categories, and expresses back the corresponding facial expressions in the same category as that of the virtual agent (infant agent). 
The facial expression recognition of the infant agent by the mother agent is based on the following rules: 1) pleasure (when the corner of the mouth is raised), 2) anger (when the corner of the mouth falls, eyebrows are knitted, and eyes are more than half open), 3) sadness (when the corner of the mouth falls, eyebrows are knitted and eyes are more than half closed), and 4) neutral (otherwise). 

This experimental design is based on a known phenomenon called ``mirroring,'' in which the mother intuitively imitates the infant's expression on a daily basis \citep{Winnicott60}.  
This is said to be important for young infants to learn emotional adjustment and social response \citep{Murray16}. 
Especially for smiling, interactive smile games between infants and their caregivers are known as an important milestone in infant social development and build the foundation for later forms of social competence \citep{Kaye80}. 
Ruvolo and colleagues revealed that there exists a strategy for the timing when the child smiles and the relationship with his/her mother \citep{Ruvolo15}. 
Thus, the purpose of this experiment is to observe the behavior learned by the infant agent and the change in interoception, emotional state, and emotion due to learning of a facial expression strategy. 

\begin{figure}[t]
\begin{center}
\includegraphics[width=150mm]{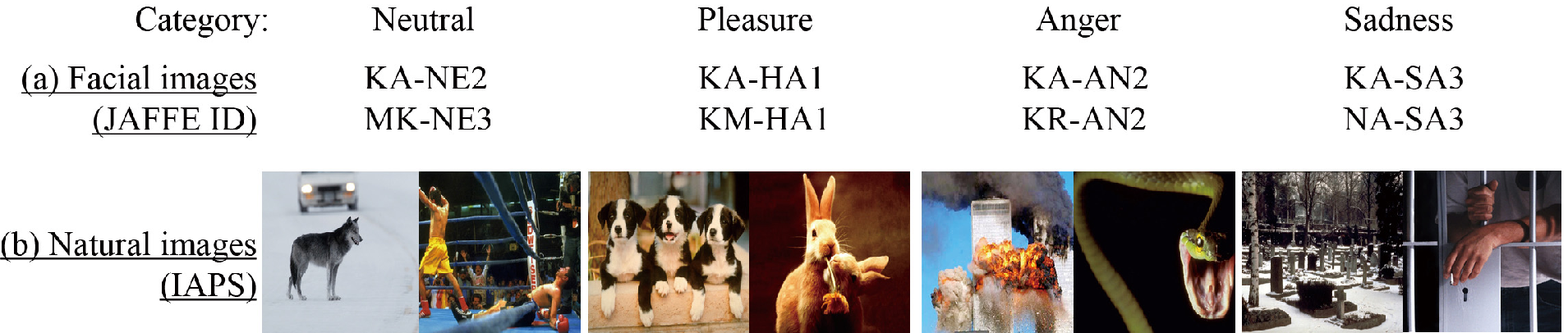}
\caption{Images used in the experiment: (a) facial images selected according to the infant agent's facial expression (JAFFE IDs are shown instead of actual images because of personality rights), and (b) natural images selected randomly.} 
\label{fig:experiment}
\end{center}
\end{figure}

In this experiment, we have two different conditions: ``face-only condition'' and ``face+natural condition.'' 
These conditions are set to compare the ideal condition of seeing only the face of the mother and the case where environmental factors exist. 
In the ``face-only'' condition, the infant agent always receives a facial image according to the infant agent expression (mirroring). 
The top row of Fig. \ref{fig:experiment} represents information on facial images, which are selected from JAFFE database \citep{Dailey2010EvidenceAA}. 
There are two different facial images for each emotional category. 
One of these two images is selected randomly to present to the infant agent. 
Although the actual images used in this experiment cannot be shown, one can check the images by downloading the database from \url{http://www.kasrl.org/jaffe.html}. 
``JAFEE ID'' corresponds to the filename of each image. 
On the contrary, in the ``face+natural'' condition, the infant agent randomly receives one of the facial images in the top row of Fig. \ref{fig:experiment} or one of the IAPS images in the bottom row of Fig. \ref{fig:experiment} as a visual stimulus. 
The natural images from IAPS mimic environmental stimuli. 
It should be noted that the facial images are stimuli that the infant agent can manipulate, because the facial images are selected according to the infant agent action. 
The IAPS images are, however, stimuli that cannot be manipulated, as they are randomly chosen. 
In other words, it is expected that the infant agent learns a policy to acquire intended stimulus according to a given facial image, and learns countermeasures, e.g., closing eyes, when an undesirable stimulus is presented from the IAPS images. 
In both cases, the image of the closed eyes portion is displayed as a black image when the agent closes his eyes. 

We performed 100000 epochs of this training using the proposed emotion model and the abovementioned scenario. 
Each time learning progresses, we visualize the middle layer of the policy network in Fig. \ref{fig:appimpl} using principal component analysis (PCA) in order to observe the state space constructed by the infant agent through the mother--infant interaction. 
If our emotions were correctly defined and properly implemented, then this state space could be divided into emotional categories by actions. 
%
%
%

\subsubsection{Results}
\begin{figure}[t]
\begin{center}
\includegraphics[width=150mm]{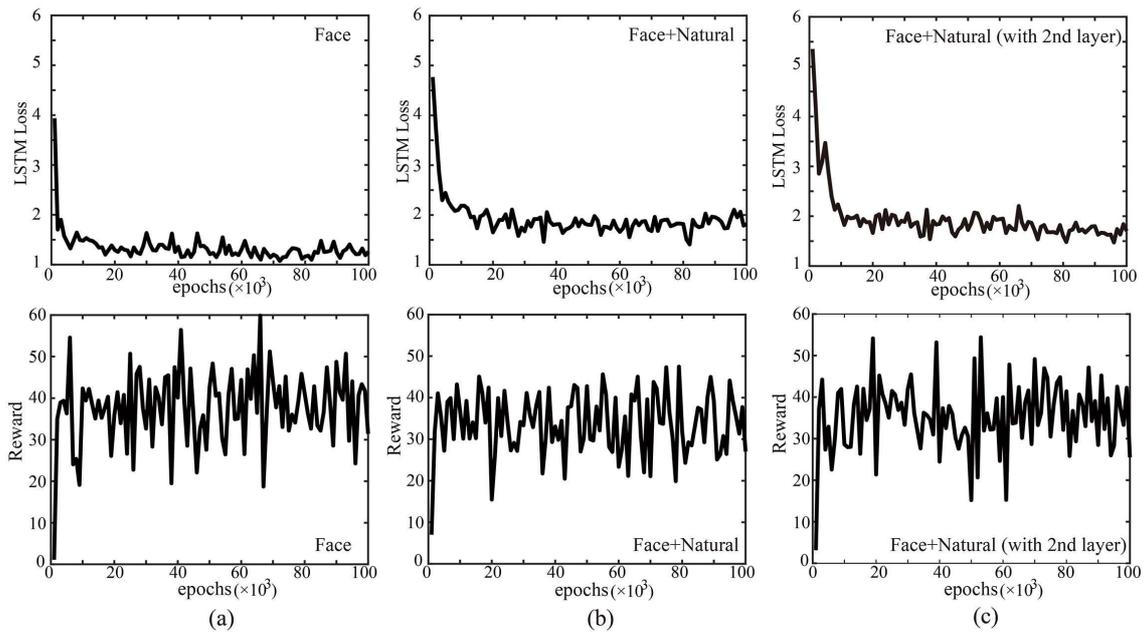}
\caption{Learning curves of the LSTM and the DDPG: (a) face-only condition in experiment 4.2, (b) face+natural condition in experiment 4.2, and (c) face+natural condition in experiment 4.3. }
\label{fig:learn1}
\end{center}
\end{figure}

Figures \ref{fig:learn1} (a) and (b) show the learning curves of this experiment for the face-only and face+natural conditions, respectively.  
On the top row, the learning curves of the LSTM are shown. 
From these figures, one can see that the LSTM learns to predict the next stimuli and interoception values. 
The training loss rapidly decreases within 5000 epochs. 
By comparing the face-only condition and face+natural condition, it is natural that the prediction error is smaller in the face-only condition. 
For the reward in the bottom row of Figs. \ref{fig:learn1} (a) and (b), similar properties can be seen. 
In fact, the reward rapidly increases for less than 5000 epochs. 
The reward does not converge to a constant value. 
This fluctuation occurs because the reward is based on homeostasis, which is a difference between a current interoception value and the past averaged interoception values. 
If there is a sudden change such as recovery of strength, the reward tends to change suddenly. 
In spite of this fluctuation of the reward, it can be clearly seen that the face-only condition achieved the higher reward in total. 
This is because the prediction in face-only condition works better than the face+natural case. 
In other words, the infant agent can better control the external environment as the mother agent always shows the facial expression in response to the infant agent. 

\begin{figure}[t]
\begin{center}
\includegraphics[width=150mm]{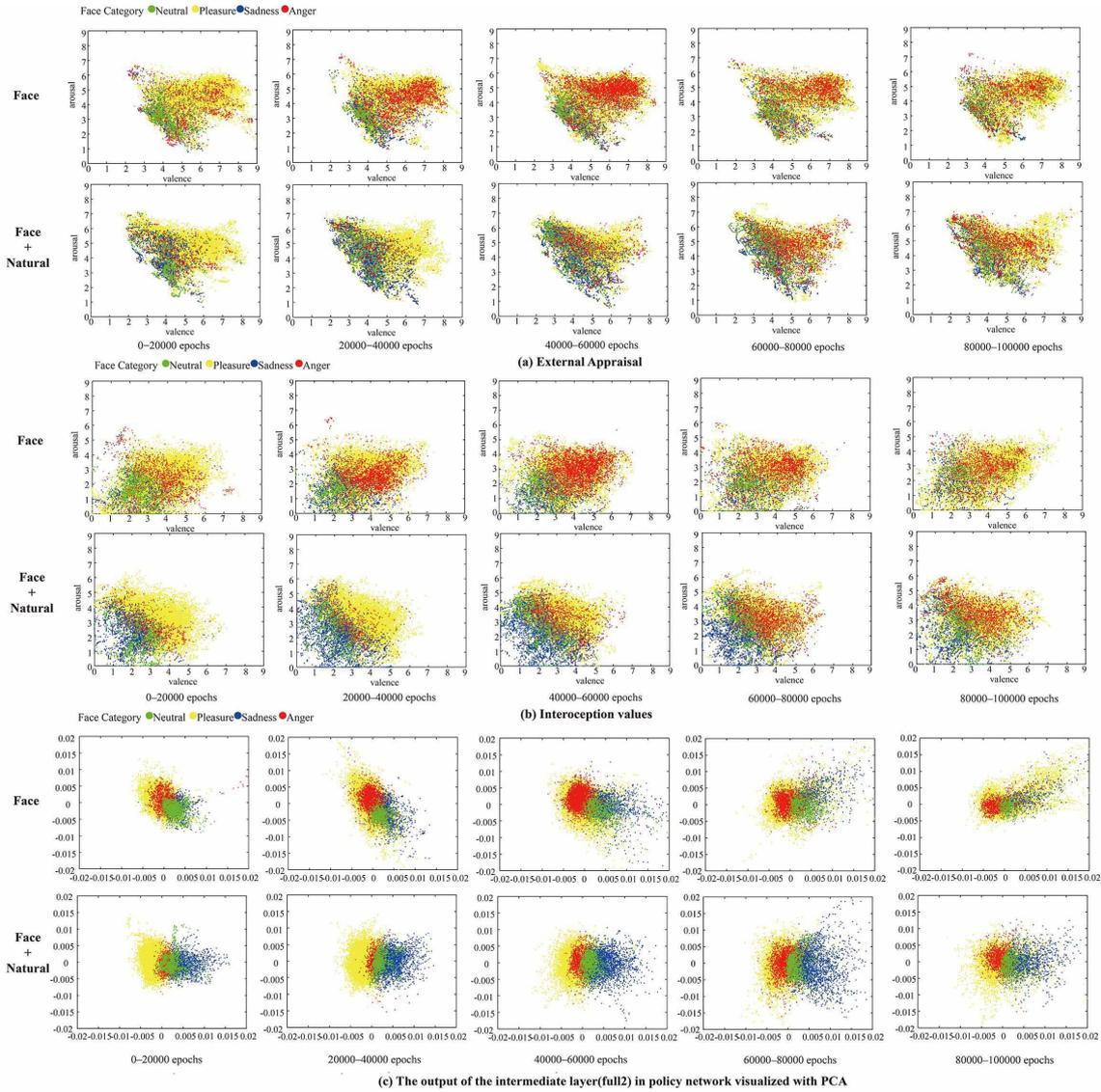}
\caption{Visualization of the internal representations in experiment 4.2 (first + third layers): (a) external appraisal during each period of epochs, (b) interoception values during each period of epochs, (c) PCA visualization of the middle layer of the policy network during each period of epochs. It should be noted that the top and bottom rows represent the results of the face-only and face+natural conditions, respectively.}
\label{fig:result}
\end{center}
\end{figure}

Now, we examine the change in internal representation, i.e., emotional state, behind this reinforcement learning.
The results are shown in Fig. \ref{fig:result}. 
Figures \ref{fig:result}(a) and (b) show plots of the external appraisal and interoception values, respectively. 
The visualization of the middle layer of the policy network using PCA is shown in Fig. \ref{fig:result} (c). 
These results correspond to (a), (b), and (c) in Fig. \ref{fig:model-impl}. 
Each color represents a category of facial expression recognized by the mother agent. 
Specifically, green, yellow, blue, and red represent neutral, pleasure, sadness, and anger, respectively. 
The top rows of Figs. \ref{fig:result} (a)--(c) show the results of the face-only condition, and the bottom rows show the results of the face+natural condition. 
As mentioned previously, the face-only condition indicates that only stimuli that can be controlled by the infant agent are provided, whereas in the face+natural condition, half of the stimuli can be controlled by the infant agent and the other half cannot. 
From the results, one can see that the colors are mixed all over in Figs. \ref{fig:result} (a) and (b). 
This implies that the external appraisal and interoception do not explicitly provide emotion differentiation functionality. 
Moreover, it can be seen that the space does not expand in Figs. \ref{fig:result} (a) and (b) as the learning progresses. 
However, the state space expands and is divided for each color as learning progresses in Fig. \ref{fig:result} (c). 
We hypothesize that this is the basic mechanism of emotion differentiation, which is observed in the middle layer of the policy network. 
Because the interoception and external appraisal did not show differentiation, these results indicate the plausibility of the proposed emotion model. 

We stop the learning process at certain epochs and run the infant agent using the learned model at each epoch to observe the behavior of the infant agent. 
From these observations, we found that the agent had the following behavioral changes (One can download the demo video of the running agent using learned models from https://youtu.be/DHOIbe4qEEY). 

In 20000 epochs model, the infant agent often opens his eyes. 
In the model of 40000 epochs, he often closes his eyes. 
He changes facial expressions by stimulation in the model of 60000 epochs. 
He closes his eyes at the times when the internal appraisal increases in the model of 80000 epochs. 
Finally, after 100000 epochs, he shows various facial expressions and has succeeded in stabilizing emotions. 
%

Essentially, the agent smiles very often and makes the other person smile. 
This behavior also seems to be altruistic, such that the agent is trying to make the partner smile. 
This behavior seems to be consistent with the findings in \citep{Ruvolo15}. 
Actually, it is interesting that the infant agent is just smiling for the desired stimulus, that is, the agent learned a selective smile. 


%
%

\subsection{Experiments on the whole system including the second layer}
\subsubsection{Experimental setup}
In Section \ref{sec:1-3}, we conducted an experiment using the first and third layers. 
This is because the integration of the first and third layers is the core part of the proposed emotion model. 
We are interested in the core mechanism of the emotion model and evaluated the model without using the second layer in the previous section. 
The whole system, including the second layer, is the focus of our interest in this section. 
Moreover, we can determine the importance of the second layer by comparing the results to the previous ones. 
We use exactly the same experimental protocol as in Section \ref{sec:1-3}; however, only the face+natural condition is adopted as it is obvious that the face-only condition gives better performance in terms of prediction. 


\subsubsection{Results}

The learning curve of the whole system is given in Fig. \ref{fig:learn1} (c). 
The upper graph represents the LSTM loss versus the number of epochs. 
This graph shows a similar tendency to the previous experiment, which means that the LSTM learns to predict the next image and interoception values. 
The lower graph shows the reward with respect to the number of epochs. 
This also shows the same tendency as the previous experiment. 

It is interesting to compare the results between the previous and current experiments in terms of average errors in the LSTM and average reward that the agent obtained. 
For the LSTM, the averaged losses, which are represented by $\ell_a$ (face-only condition without second layer), $\ell_b$ (face+natural condition without second layer), and $\ell_c$ (face+natural condition with second layer) are expected to be in the order $\ell_{a}$ $<$ $\ell_{c}$ $<$ $\ell_{b}$. 
In fact, the averaged losses are $\ell_{a}= 1.26$, $\ell_{b}=1.89$, and $\ell_{c}=1.66$; and the order of these values is as expected. 
Exactly the same observations can be made with respect to the reward (larger is better in this case). 
The averaged reward values are $\bar{R}_{a}= 38.72$, $\bar{R}_{b}=32.35$, and $\bar{R}_{c}=35.24$ ($\bar{R}_a > \bar{R}_c > \bar{R}_b$). 
These results are obtained because the face-only condition is the easiest setting for the infant agent. 
Moreover, the second layer improves the prediction of the next situation, which leads to $\bar{R}_{c} > \bar{R}_{b}$. 

%
In order to show that the second layer actually works, the learned models (both with and without the second layer) were run for 3000 epochs and the interoception values were collected. 
Then, the mean absolute differences (MAD) $|\bm{a}_t -\bm{a}_{t+1}|$ of both models were compared. 
For the valence, the MAD of the previous experiment (face+natural condition without second layer) was $0.21$. 
The MAD of the current experiment (face+natural condition with second layer) was $0.16$. 
For the arousal, the MAD of the previous experiment and the current experiment were, respectively, $0.20$ and $0.15$. 
The t-test was performed on the MAD of both models and revealed that the MAD was significantly smaller in the case with the second layer ($p < 0.01$). 
This indicates that the second layer works as we expected, and it improves the learning performance. 

\begin{figure}[t]
\begin{center}
\includegraphics[width=150mm]{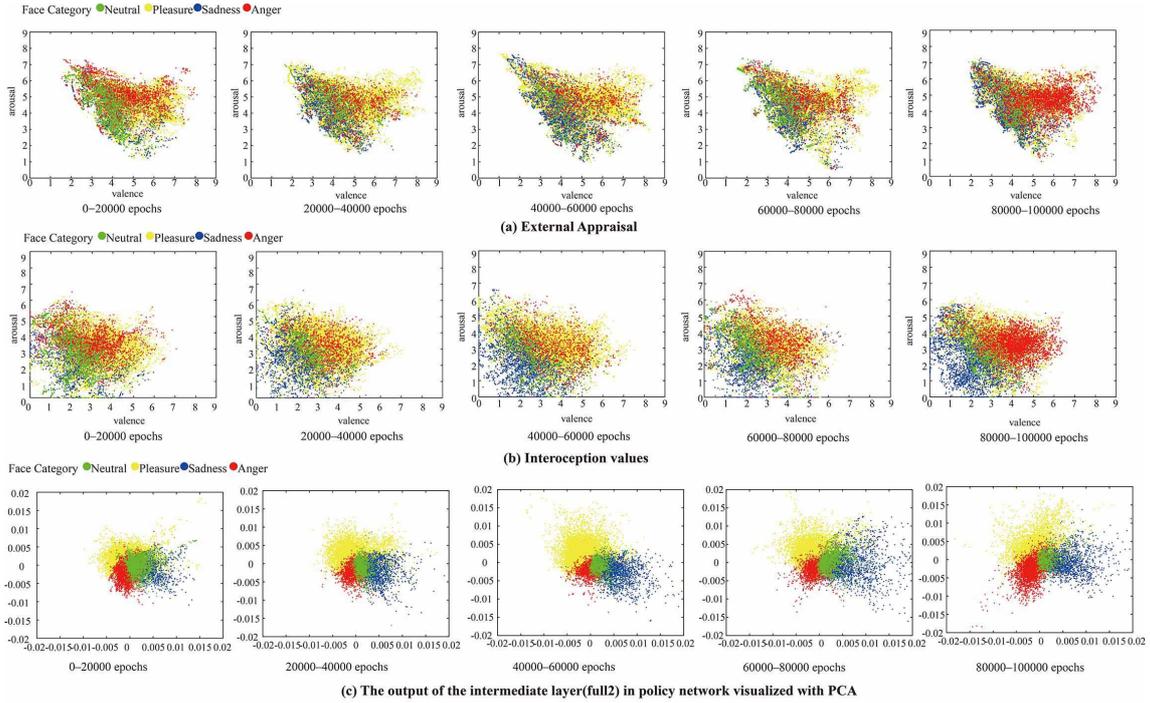}
\caption{Visualization of the internal representations in experiment 4.3 (the whole model): (a) external appraisal, (b) interoception, and (c) PCA visualization of the middle layer of the policy network during each period of epochs.}
\label{fig:result-123}
\end{center}
\end{figure}

Here, discuss in detail the representation inside the network to find the basis of emotions. 
Figure \ref{fig:result-123} shows plots of the external appraisal, interoception values, and visualization of the middle layer of the policy network using PCA. 
These results correspond to (a), (b), and (c) in Fig. \ref{fig:model-impl}. 
Each color represents a category of facial expression recognized by the mother agent, as mentioned in the previous section. 
From these figures, one can see that the representation in the policy network divides the emotional category very clearly compared with the external appraisal and the interoception. 
Moreover, it is also clear that the policy network represents categories far better compared with Fig. \ref{fig:result}, which does not include the second layer. 

\begin{figure}[t]
\begin{center}
\includegraphics[width=150mm]{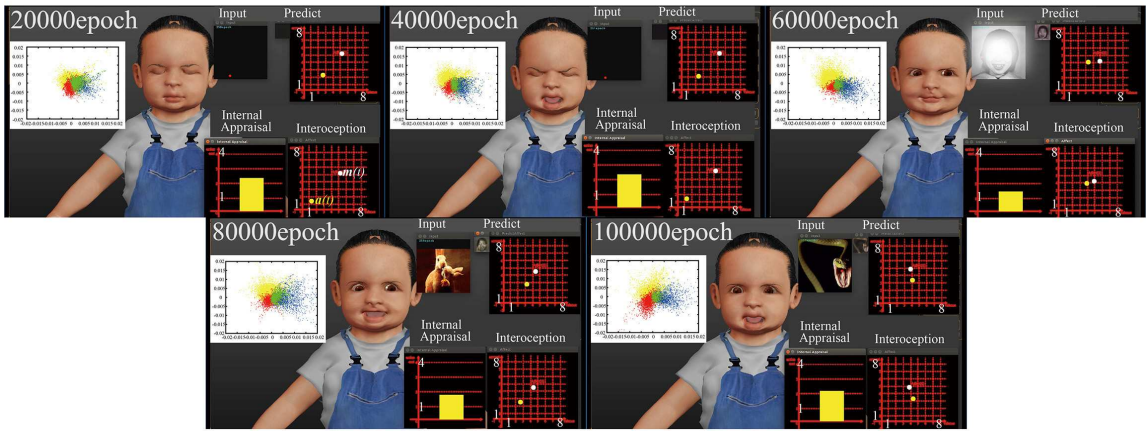}
\caption{Examples of facial expressions by the infant agent using the learned model with the second layer. Please note that the facial input image on top right is blurred for personality rights. }
\label{fig:face}
\end{center}
\end{figure}

In this experiment, we also run the agent using the learned model. 
Figure \ref{fig:face} shows some typical facial expressions by the infant agent for each model at specific epochs. 
From the observations of the infant agent's behavior, we found that the agent had the following behavioral changes (One can download the demo video of the running agent using learned models from https://youtu.be/Phjn58kJ2ns): \\
{\bf --20000 epochs:} The agent often closes his eyes. \\
{\bf --40000 epochs:} The agent often closes his eyes. \\
{\bf --60000 epochs:} The agent often opens his eyes and he changes expressions by stimulation. \\
{\bf --80000 epochs:} The agent closes his eyes at the times when the internal appraisal increases. \\
{\bf --100000 epoch:} The agent shows various facial expressions (surprise, anger, etc.) and has succeeded in stabilizing affects.\\

%
%

%
%

%
%

\begin{figure}[t]
\begin{center}
\includegraphics[width=100mm]{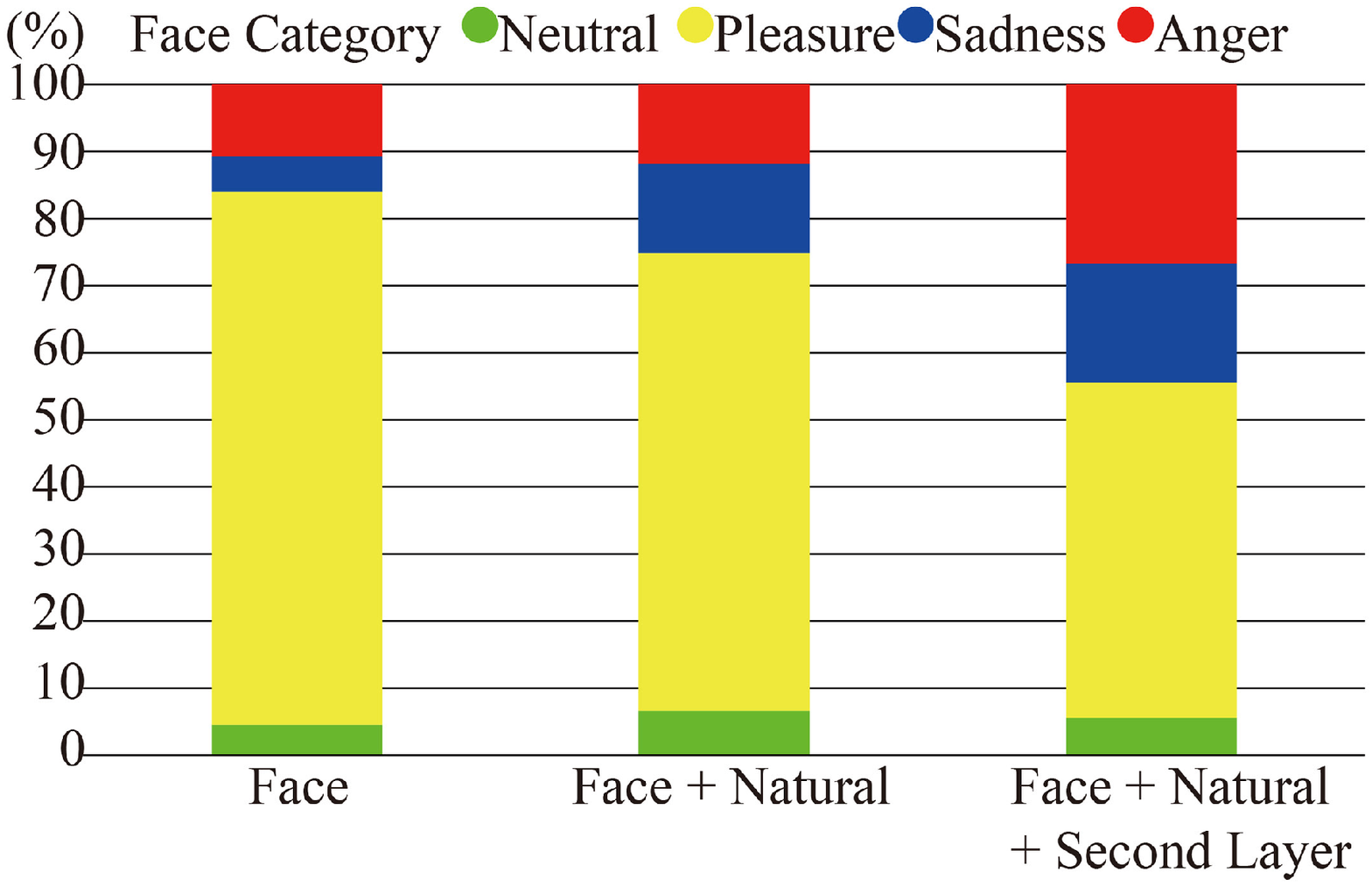}
\caption{Frequency ratio of facial expressions for each condition.}
\label{fig:action-bar}
\end{center}
\end{figure}

\subsection{Discussion}
In the first experiment, the RAM was evaluated. 
The results of this experiment show that the RAM has an ability to replicate the innate reactions of a human against specific stimulation. 
It is interesting that although the network does not learn the reactions directly, it can learn general human reactions. 
For example, when an image of a pleasure facial expression is input to the RAM, the arousal and valence values corresponding to pleasure are generated. 
Moreover, similar responses of infants to color are learned by the RAM. 
These facts indicate the existence of an innate and general response of humans to visual stimuli, and the RAM can extract such visual features. 
For the implementation of emotional robots, emotions are usually designed manually by the robot designer and the above results may free the robot designer from this difficult design task. 

In the second experiment, we evaluated the proposed emotion models. 
According to the PCA results with the face-only condition, pleasure occupies half, and the remaining half seems to consist of a mix of neutral and sadness, and anger can be seen to a lesser degree. 
In an environment that the agent can always control, anger is not necessary to be output; that is, the agent learned to deal with stimuli by pleasure or otherwise. 
However, according to the result of the face+natural condition, although pleasure is predominant, anger increases, and neutral and sadness are separated as compared with the face-only condition. 
This is due to the necessity of selecting actions by classifying stimuli in more detail because of uncontrollable stimuli. 
Therefore, it can be surmised that not only the controllable stimuli but also the uncontrollable stimuli create our human-like rich emotions. 
The uncontrollable stimuli also give a very important meaning to learning to predict the future; that is, if the world is simple enough to predict perfectly, then the learning does not mean anything. 

In the third experiment, the whole emotion model including the second layer was evaluated. 
By having the second layer in the emotion model, the state space, i.e., the middle layer of the policy network, has the representations of basic emotions such as anger, sadness, pleasure, and neutral. 
More interestingly, these emotional categories are located as assumed in the dimensional model; that is, neutral is located at the center, and pleasure, sad, and anger are located surrounding the neutral. 
Pleasure does not occupy the PCA space anymore, and it seems to be relatively evenly divided. 
In particular, the frequency of anger increases as shown in Fig. \ref{fig:action-bar}. 
Because the second layer works as a smoothing function, the interoception values of temporally adjacent stimuli are made closer, and sudden changes are reduced. 
As a result, prediction in the LSTM improves, and categorization of the stimulus is promoted. 
It is thought that these effects result in relatively uniform and distinct differentiation of the boundary surface of emotional categories. 

Now, let us consider the behavioral output of the infant agent with the whole emotion model. 
In the early stage of learning, the agent closed his eyes well and the eyes are opened well in the second half of the early stage. 
This is similar to the development of infants. 
In general, infants initially almost always have their eyes closed (sleeping), and the time with their eyes open increases gradually. 
This process may be mainly dependent on the developmental process of the physical bodies of young infants. 
However, in the course of action selection, infants may have a stage to learn that the best policy is to close the eyes at the beginning, 
and gradually shift toward the policy of keeping their eyes open. 
This is only a speculation, which should be verified in the future. 
Additionally, the 100000 epoch result in Fig. \ref{fig:face} shows that the infant agent looks surprised by the snake. 
In the PCA space of the middle layer of the policy network, i.e., internal representation of emotional states, it is not clear whether the surprise category was generated, because the actions were classified with only four emotional categories. 
However, there is a possibility that a richer emotional space emerged as the internal representation of the proposed emotion model. 
This point still needs further analysis. 

Here, the limitations of our proposed model are discussed. 
Because the IAPS used adult human subjects to label the arousal and valence values, there must be an issue of the RAM using the IAPS in the first place. 
However, we think that the averaging process of the labeled values reduced the individuality of the data and innate reactions were extracted. 
The results of the first experiment using the RAM implies that this is in fact true. 
Currently, biosignals from a real human body instead of the IAPS database are prepared to use for training the RAM as another direction of this research. 
Another issue to be addressed is the reward for reinforcement learning, which is currently based solely on the idea of ``homeostasis.''
The idea of intrinsic motivation that appeared as a series of counterarguments to drive reduction theory cannot be ignored \citep{kage1994_ja}. 
More complex tasks should be considered in the future, because the current facial expression task is too simple to examine the full functionality of the emotion model. 
We also consider using a real robot to examine more complex internal appraisals. 

From the viewpoint of empathic communication, ``other'' should appear in Fig.\ref{fig:model-impl}. 
Moreover, self/other discrimination must be considered in the model for generating higher-level social emotions. 
Language is another important aspect of the emotion model \citep{Lieberman2007PuttingFI}. 
We are currently working on the ``emotional symbol grounding problem'' using the idea of language acquisition by robots \citep{nagai2017}. 
%
In addition, it is necessary to consider empathy. 
For example, Lim et al. proposed multimodal emotional intelligence \citep{lim2015recipe}. 
Their model was inspired by the mirror neuron system, which is a mechanism underlying human cognition \citep{iacoboni2009imitation}. 
In considering empathy, the work on mirror neurons cannot be ignored.

%% file: conclusion.tex
%
\section{Conclusions}
In this study, a computational model of emotion, which consists of three layers was proposed. 
As the first layer, we examined a method for generating valence and arousal values by given visual stimuli using the RAM. 
Some promising results were obtained, which verified that the first layer is plausible for generating human-like quick reactions against specific stimuli. 
Next, we examined a decision-making mechanism, which is the third layer, by employing a convolutional LSTM and DDPG. 
As a result, the agent learned a selective smile and emotion differentiation was observed. 
Finally, the whole model including the second layer was integrated and its performance was studied. 
The results obtained in this experiment show that the second layer provided far better results compared with the model without the second layer. 
For future work, we will evaluate the proposed model using more complex tasks. 
The implementation on a real physical robot is also left for future work.

%% file: appendix.tex
\appendix
\section*{Appendices}
\setcounter{section}{0} 
\renewcommand{\thesection}{\Alph{section}} 
\setcounter{equation}{0} 
\renewcommand{\theequation}{\Alph{section}.\arabic{equation}}
\setcounter{figure}{0} 
\renewcommand{\thefigure}{\Alph{section}.\arabic{figure}}
\setcounter{table}{0} 
\renewcommand{\thetable}{\Alph{section}.\arabic{table}}
\section{Recurrent attention model (RAM)}

The RAM is a recurrent neural network (RNN) with visual attention proposed by Mnih et al. \citep{RAM2014}. 
In general, humans focus attention selectively on parts of the visual space instead of processing whole scene at once. 
Human visual perception acquires information when and where it is needed, and combine information from different fixations over time. 
This is how we build up an internal representation of the scene and we use the representation for decision making. 
Based on this idea the RAM, which is a novel framework for attention-based task-driven visual processing with neural networks, has been developed. 

As shown in Fig. \ref {fig:model-ram} (b), images with multiple resolutions are acquired from the original image $x_{t}$ at the center point $l_{t-1}$. 
Then, each point and multiple images are input to the linear layer as $g_{t} = f_{g}(x_{t},l_{t-1};\theta_{g})$. 
$f_{h}(\theta_{h})$ is the core network and takes $h_{t-1}$, which is a previous internal representation, as an input. 
The action network $f_ {a}(\theta_{a})$ and the location network $f_{l}(\theta_{l})$ take $h_t$ to calculate the valence/arousal values and location of the next step, respectively. 

The parameters of RAM are defined as $\theta = \{ \theta_{g}, \theta_{h}, \theta_{a} \}$, and $\theta$ is optimized such that the total reward the agent can obtain when interacting with the environment is maximized. More specifically, the policy of the agent induces a distribution over possible interaction sequences $s_{1:N}$ and the reward is maximize under this distribution: 
\begin{equation}
J(\theta)={\bm E}_{p(s_{1:T};\theta)}[\sum_{t=1}^{T}r_{t}]={\bm E}_{p(s_{1:T};\theta)}[R], 
\end{equation}
where $p(s_{1:T};\theta)$ depends on the policy. 
Although it is difficult to maximize $J$ exactly, we can apply some techniques form the reinforcement learning by viewing the problem as a partially observable Markov decision process.  
In this case, the gradient can be expressed as  
\begin{equation}
  \nabla_{\theta}J=\sum_{t=1}^{T} {\bm E}_{p(s_{1:T};\theta)}[\nabla_{\theta}log\pi(u_{t} | s_{1:t};\theta)R] \approx \frac{1}{M}\sum_{i=1}^{M}\sum_{t=1}^{T}\nabla_{\theta}log\pi(u_{t}^{i} | s_{1:t}^{i};\theta)R^{i},
\label{equ:ram}
\end{equation}
where $s^{i}$ are interaction sequences obtained by running the current agent $\pi_{\theta}$ for $i = 1 \cdots M$ episodes. 
The learning rule is also known as the REINFORCE rule. 
It involves running the agent with its current policy to obtain samples of interaction sequences $s_{1:T}$. 
Then, the parameters $\theta$ of the agent are adjusted such that the log-probability of the chosen actions that have led to high cumulative reward is increased, while that of actions having produced low reward is decreased. 
Eq. (\ref{equ:ram}) requires us to compute $\nabla_{\theta}log\pi(u_{t} | s_{1:t};\theta)$; however, this is the gradient of the RNN that defines the agent evaluated at time step $t$ and can be computed by standard backpropagation. 

In our scenario, the RAM must output the arousal/valence values for the input image as the final action. 
For the training images, these values are known and the policy, that outputs the correct values associated with a training image at the end of an observation sequence, can be directly optimized.  
This can be achieved by maximizing the conditional probability of the true values given the observations from the image, i.e., by maximizing $log\pi(a_{T}^{*}|s_{1:T};\theta)$, where $a_{T}^{*}$ corresponds to the ground-truth associated with the image from which observations $s_{1:T}$ were obtained. 
The original RAM follows this approach for classification problems, where it optimizes the cross-entropy loss to train the action network $f_{a}$ and the gradients are backpropagated through the core and glimpse networks. 
The location network $f_{l}$ is always trained with REINFORCE, which provides the parameter $\theta_l$. 
\section{Convolutional long short-term memory (LSTM)}
Convolutional LSTM is a method combining CNN, which captures the features of images, and LSTM, which can handle long-term time series information, proposed by Xingjian et al. \citep{xingjian2015convolutional}. 
Specifically, it is a network in which multiplication by the weight of LSTM is convolution, and the constituent element is composed of a memory cell $C_{t}$, input gate $i_{t}$, forget gate $f_{t}$, and output gate $o_{t}$. 
\begin{eqnarray}
 i_{t}&=&\sigma(W_{xi}*X_{t}+W_{hi}*H_{t-1}+W_{ci}\circ C_{t-1}+b_{i}), \\
 f_{t}&=&\sigma(W_{xf}*X_{t}+W_{hf}*H_{t-1}+W_{cf}\circ C_{t-1}+b_{f}), \\
 C_{t}&=&f_{t}\circ C_{t-1}+i_{t}\circ tanh(W_{xc}*X_{t}+W_{hc}*H_{t-1}+b_{c}), \\
 o_{t}&=&\sigma(W_{xo}*X_{t}+W_{ho}*H_{t-1}+W_{co}\circ C_{t}+b_{o}), \\
 H_{t}&=&o_{t}\circ tanh(C_{t}),
\end{eqnarray}
where $X_{t}$ are inputs, $H_{t}$ are hidden states, the $W$ terms denote weight matrices, the $b$ terms denote bias vectors, $*$ denotes the convolution operator, and $\circ$ denotes the Hadamard product. 

The memory cells are responsible for storing past states. The input gate has a role of adjusting the value added to the memory cell. It is possible to prevent the important information possessed by the memory cell from being lost due to the influence of the most unrelated information that is most recent, owing to the existence of this gate. The forget gate has a role of adjusting how much the value of the memory cell is held at the next time. The output gate serves to adjust how much the value of the memory cell affects the next layer. The existence of this gate can prevent the entire network from being disturbed by short-term memory and interruption of long-term memory. 

In this study, we use two layers of convolutional LSTM; the filter is $5 \times 5 \times 5$ and the error is calculated by the mean square error. The learning rate is adaptive moment estimation (Adam) ($\alpha=0.001,\beta_{1}=0.9,\beta_{2}=0.999,\epsilon=10^{-8}$).

\section{Deep deterministic policy gradient (DDPG)}
DDPG is a reinforcement learning method using deep learning proposed by Lillicrap et al. \citep{DDPG}.
As recently reported, ``Deep Q Network” (DQN) algorithm \citep{mnih2015human} is capable of human-level performance on many Atari video games using unprocessed pixels for input. 
Whereas DQN solves problems with high-dimensional observation spaces, it can only handle discrete and low-dimensional action spaces.
Then, they presented a model-free, off-policy actor-critic algorithm (DDPG) using deep function approximators that can learn policies in high-dimensional, continuous action spaces. 
The DDPG algorithm is shown in Algorithm \ref{alg1}.
The learning rate is Adam (actor network: $\alpha=10^{-4},\beta_{1}=0.9,\beta_{2}=0.999,\epsilon=10^{-8}$, critic network: $\alpha=10^{-3},\beta_{1}=0.9,\beta_{2}=0.999,\epsilon=10^{-8}$).
$\mathcal{N}$ is the Ornstein--Uhlenbeck process. $N_B$ is 200. The size of $B$ is 500. When new data comes in, old data is discarded.
We used batch normalization.
\begin{algorithm}                      
\caption{DDPG algorithm}         
\label{alg1}                          
\begin{algorithmic}                  
\STATE Randomly initialize critic network $Q(s,a|\theta^{Q})$ and actor $\mu(s|\theta^{\mu})$ with weights $\theta^{Q}$ and $\theta^{\mu}$. 
\STATE Initialize target network $Q'$ and $\mu '$ with weights $\theta^{Q'} \leftarrow \theta^{Q}, \theta^{\mu '} \leftarrow \theta^{\mu}$
\STATE Initialize replay buffer $B$
\FOR{episode = 1, M}
\STATE Initialize a random process $\mathcal{N}$ for action exploration
\STATE Receive initial observation state $s_{1}$
\FOR{t = 1, T}
\STATE Select action $a_{t} = \mu(s_{t}|\theta^{\mu}) + \mathcal{N}_{t} $ according to the current policy and exploration noise
\STATE Execute action $a_{t}$ and observe reward $r_{t}$ and observe new state $s_{t+1}$
\STATE Store transition ($s_{t}, a_{t}, r_{t}, s_{t+1}$) in $B$
\STATE Sample a random minibatch of $N_B$ transitions ($s_{i}, a_{i}, r_{i}, s_{i+1}$) from $B$
\STATE Set $y_{i}=r_{i}+\gamma Q'\left (s_{i+1},\mu '(s_{i+1}|\theta^{\mu '})|\theta^{Q'} \right) $
\STATE Update critic by minimizing the loss: $L=\frac{1}{N_B}\sum_{i} \left \{ y_{i}-Q(s_{i},a_{i}|\theta^{Q}) \right \}^{2}$
\STATE Update the actor policy using the sampled policy gradient:\\ $\nabla_{\theta^{\mu}}J \approx \frac{1}{N_B}\sum_{i}\nabla_{a}Q(s,a|\theta^{Q})|_{s=s_{i},a=\mu(s_{i})}\nabla_{\theta~{\mu}} \mu(s|\theta^{\mu})_{s_{i}}$
\STATE Update the target networks:\\ $\theta^{Q'} \leftarrow \eta \theta^{Q}+(1-\eta)\theta^{Q'}$\\ $\theta^{\mu'} \leftarrow \eta \theta^{\mu}+(1-\eta)\theta^{\mu'}$
\ENDFOR
\ENDFOR
\end{algorithmic}
\end{algorithm}

%% file: main.bbl
\begin{thebibliography}{66}
\providecommand{\natexlab}[1]{#1}
\expandafter\ifx\csname urlstyle\endcsname\relax
  \providecommand{\doi}[1]{doi:\discretionary{}{}{}#1}\else
  \providecommand{\doi}{doi:\discretionary{}{}{}\begingroup
  \urlstyle{rm}\Url}\fi
\providecommand{\selectlanguage}[1]{\relax}
\providecommand{\bibAnnoteFile}[1]{%
  \IfFileExists{#1}{\begin{quotation}\noindent\textsc{Key:} #1\\
  \textsc{Annotation:}\ \input{#1}\end{quotation}}{}}
\providecommand{\bibAnnote}[2]{%
  \begin{quotation}\noindent\textsc{Key:} #1\\
  \textsc{Annotation:}\ #2\end{quotation}}

\bibitem[{Arnold(1960)}]{Arnold1960}
Arnold, M.~B. (1960).
\newblock \emph{Emotion and Personality}.
\newblock Emotion and Personality (Cassell \& Company)
\bibAnnoteFile{Arnold1960}

\bibitem[{Asada(2015)}]{ASADA15}
Asada, M. (2015).
\newblock Development of artificial empathy.
\newblock \emph{Neuroscience Research} 90, 41--50.
\newblock \doi{https://doi.org/10.1016/j.neures.2014.12.002}
\bibAnnoteFile{ASADA15}

\bibitem[{Barrett et~al.(2015)Barrett, Feldman, and Simmons}]{EPIC2015}
Barrett, Feldman, L., and Simmons, W.~K. (2015).
\newblock Interoceptive predictions in the brain.
\newblock \emph{Nature reviews. Neuroscience} 16.7, 419--429
\bibAnnoteFile{EPIC2015}

\bibitem[{Barsade(2002)}]{Barsade02}
Barsade, S.~G. (2002).
\newblock The ripple effect: Emotional contagion and its influence on group
  behavior.
\newblock \emph{Administrative Science Quarterly} 47, 644--675.
\newblock \doi{10.2307/3094912}
\bibAnnoteFile{Barsade02}

\bibitem[{Breazeal(2002)}]{DesigningSociable}
Breazeal, C. (2002).
\newblock Designing sociable robots.
\newblock \emph{The MIT Press}
\bibAnnoteFile{DesigningSociable}

\bibitem[{Bridges(1932)}]{Bridges}
Bridges, K. M.~B. (1932).
\newblock Emotional development in early infancy.
\newblock \emph{Child development} , 324--341
\bibAnnoteFile{Bridges}

\bibitem[{Ca{\~n}amero and Gaussier(2005)}]{canamero2005emotion}
Ca{\~n}amero, L. and Gaussier, P. (2005).
\newblock Emotion understanding: robots as tools and models.
\newblock \emph{Emotional Development} , 235--258
\bibAnnoteFile{canamero2005emotion}

\bibitem[{Cannon(1927)}]{Cannon}
Cannon, W.~B. (1927).
\newblock The james-lange theory of emotions: A critical examination and an
  alternative theory 39, 106^^e2^^80^^93--124
\bibAnnoteFile{Cannon}

\bibitem[{Dailey et~al.(2010)Dailey, Joyce, Lyons, Kamachi, Ishi, Gyoba
  et~al.}]{Dailey2010EvidenceAA}
Dailey, M.~N., Joyce, C., Lyons, M.~J., Kamachi, M., Ishi, H., Gyoba, J.,
  et~al. (2010).
\newblock Evidence and a computational explanation of cultural differences in
  facial expression recognition.
\newblock \emph{Emotion} 10 6, 874--93
\bibAnnoteFile{Dailey2010EvidenceAA}

\bibitem[{Damashio et~al.(1996)Damashio, Everitt, and Bishop}]{Damashio}
Damashio, A.~R., Everitt, B.~J., and Bishop, D. (1996).
\newblock The somatic marker hypothesis and the possible functions of the
  prefrontal cortex [and discussion].
\newblock \emph{Philosophical Transactions of the Royal Society B, Biological
  Sciences} 351, 1413--1420
\bibAnnoteFile{Damashio}

\bibitem[{Damasio(2003)}]{damasio2003looking}
Damasio, A. (2003).
\newblock \emph{Looking for Spinoza: Joy, Sorrow, and the Feeling Brain}.
\newblock Harvest books (Harcourt)
\bibAnnoteFile{damasio2003looking}

\bibitem[{Dan-Glauser and Scherer(2011)}]{dan2011geneva}
Dan-Glauser, E.~S. and Scherer, K.~R. (2011).
\newblock The geneva affective picture database (gaped): a new 730-picture
  database focusing on valence and normative significance.
\newblock \emph{Behavior research methods} 43, 468
\bibAnnoteFile{dan2011geneva}

\bibitem[{Dutton(1974)}]{Dutton}
Dutton, D.~G. (1974).
\newblock Some evidence for heightened sexual attraction under conditions of
  high anxiety.
\newblock \emph{Journal of Personality and Social Psychology} 30, 510--517
\bibAnnoteFile{Dutton}

\bibitem[{Ekman and Wallace(1971)}]{ekman}
Ekman, P. and Wallace, F.~V. (1971).
\newblock Constants across cultures in the face and emotion.
\newblock \emph{Journal of personality and social psychology} 17, 124--129
\bibAnnoteFile{ekman}

\bibitem[{Friston et~al.(2009)Friston, Daunizeau, and Kiebel}]{Friston2009PLOS}
Friston, K.~J., Daunizeau, J., and Kiebel, S.~J. (2009).
\newblock Reinforcement learning or active inference?
\newblock \emph{PLOS ONE} 4, 1--13.
\newblock \doi{10.1371/journal.pone.0006421}
\bibAnnoteFile{Friston2009PLOS}

\bibitem[{Friston et~al.(2010)Friston, Daunizeau, Kilner, and
  Kiebel}]{Friston2010}
Friston, K.~J., Daunizeau, J., Kilner, J., and Kiebel, S.~J. (2010).
\newblock Action and behavior: a free-energy formulation.
\newblock \emph{Biological Cybernetics} 102, 227--260.
\newblock \doi{10.1007/s00422-010-0364-z}
\bibAnnoteFile{Friston2010}

\bibitem[{Friston and Stephan(2007)}]{Friston2007}
Friston, K.~J. and Stephan, K.~E. (2007).
\newblock Free-energy and the brain.
\newblock \emph{Synthese} 159, 417--458.
\newblock \doi{10.1007/s11229-007-9237-y}
\bibAnnoteFile{Friston2007}

\bibitem[{Hatfield et~al.(1993)Hatfield, Cacioppo, and Rapson}]{Hatfield93}
Hatfield, E., Cacioppo, J.~T., and Rapson, R.~L. (1993).
\newblock Emotional contagion.
\newblock \emph{Current Directions in Psychological Science} 2, 96--100.
\newblock \doi{10.1111/1467-8721.ep10770953}
\bibAnnoteFile{Hatfield93}

\bibitem[{Hieida et~al.(2018{\natexlab{a}})Hieida, Horii, and
  Nagai}]{hieida2018decision}
Hieida, C., Horii, T., and Nagai, T. (2018{\natexlab{a}}).
\newblock Decision-making in emotion model.
\newblock In \emph{Companion of the 2018 ACM/IEEE International Conference on
  Human-Robot Interaction}. 127--128
\bibAnnoteFile{hieida2018decision}

\bibitem[{Hieida et~al.(2018{\natexlab{b}})Hieida, Horii, and
  Nagai}]{hieida2018ROMAN}
Hieida, C., Horii, T., and Nagai, T. (2018{\natexlab{b}}).
\newblock Emotion differentiation based on decision-making in emotion model.
\newblock In \emph{IEEE International Conference on Robot and Human Interactive
  Communication}. to appear
\bibAnnoteFile{hieida2018ROMAN}

\bibitem[{Hieida and Nagai(2017)}]{HRI2017}
Hieida, C. and Nagai, T. (2017).
\newblock A model of emotion for empathic communication.
\newblock \emph{Companion of the 2017 ACM/IEEE International Conference on
  Human-Robot Interaction} , 133--134
\bibAnnoteFile{HRI2017}

\bibitem[{Iacoboni(2009)}]{iacoboni2009imitation}
Iacoboni, M. (2009).
\newblock Imitation, empathy, and mirror neurons.
\newblock \emph{Annual review of psychology} 60, 653--670
\bibAnnoteFile{iacoboni2009imitation}

\bibitem[{Izard.(1977)}]{izard1977}
Izard., C.~E. (1977).
\newblock \emph{Human emotions} (Springer US)
\bibAnnoteFile{izard1977}

\bibitem[{James(1884)}]{james}
James, W. (1884).
\newblock What is an emotion ?
\newblock \emph{Mind} os-IX, 188--205.
\newblock \doi{10.1093/mind/os-IX.34.188}
\bibAnnoteFile{james}

\bibitem[{Kage(1994)}]{kage1994_ja}
Kage, M. (1994).
\newblock A critical review of studies on intrinsic motivation.
\newblock \emph{Japanese Journal of Educational Psychology} 42, 345--359
\bibAnnoteFile{kage1994_ja}

\bibitem[{Kaye and Fogel(1980)}]{Kaye80}
Kaye, K. and Fogel, A. (1980).
\newblock The temporal structure of face-to-face communication between mothers
  and infants 16, 454--464
\bibAnnoteFile{Kaye80}

\bibitem[{Koelsch et~al.(2015)Koelsch, Jacobs, Menninghaus, Liebal,
  Klann-Delius, von Scheve et~al.}]{KOELSCH20151}
Koelsch, S., Jacobs, A.~M., Menninghaus, W., Liebal, K., Klann-Delius, G., von
  Scheve, C., et~al. (2015).
\newblock The quartet theory of human emotions: An integrative and
  neurofunctional model.
\newblock \emph{Physics of Life Reviews} 13, 1--27.
\newblock \doi{https://doi.org/10.1016/j.plrev.2015.03.001}
\bibAnnoteFile{KOELSCH20151}

\bibitem[{Kurdi et~al.(2017)Kurdi, Lozano, and Banaji}]{kurdi2017introducing}
Kurdi, B., Lozano, S., and Banaji, M.~R. (2017).
\newblock Introducing the open affective standardized image set (oasis).
\newblock \emph{Behavior research methods} 49, 457--470
\bibAnnoteFile{kurdi2017introducing}

\bibitem[{Lang et~al.(1999{\natexlab{a}})Lang, Bradley, and Cuthbert}]{IAPS}
Lang, P.~J., Bradley, M.~M., and Cuthbert, B.~N. (1999{\natexlab{a}}).
\newblock International affective picture system (iaps): Technical manual and
  affective ratings.
\newblock \emph{Gainesville, FL: The Center for Research in Psychophysiology,
  University of Florida}
\bibAnnoteFile{IAPS}

\bibitem[{Lang et~al.(1999{\natexlab{b}})Lang, Bradley, Cuthbert
  et~al.}]{lang1999international}
Lang, P.~J., Bradley, M.~M., Cuthbert, B.~N., et~al. (1999{\natexlab{b}}).
\newblock International affective picture system (iaps): Instruction manual and
  affective ratings.
\newblock \emph{The center for research in psychophysiology, University of
  Florida}
\bibAnnoteFile{lang1999international}

\bibitem[{Lazarus(1991)}]{SLazarus1991}
Lazarus, R.~S. (1991).
\newblock \emph{Emotion and Adaptation} (Oxford University Press USA)
\bibAnnoteFile{SLazarus1991}

\bibitem[{LeDoux(1986)}]{LeDoux1986}
LeDoux, J.~E. (1986).
\newblock \emph{Neurobiology of emotion} (Cambridge University Press)
\bibAnnoteFile{LeDoux1986}

\bibitem[{LeDoux(1989)}]{LeDoux1989}
LeDoux, J.~E. (1989).
\newblock Cognitive-emotional interactions in the brain.
\newblock \emph{Cognition and Emotion} 3, 267--289.
\newblock \doi{10.1080/02699938908412709}
\bibAnnoteFile{LeDoux1989}

\bibitem[{LeDoux(1998)}]{ledoux1998emotional}
LeDoux, J.~E. (1998).
\newblock \emph{The Emotional Brain: The Mysterious Underpinnings of Emotional
  Life}.
\newblock A Touchstone book (Simon \& Schuster)
\bibAnnoteFile{ledoux1998emotional}

\bibitem[{Ledoux(1998)}]{Ledoux}
Ledoux, J.~E. (1998).
\newblock The emotional brain: The mysterious underpinnings of emotional life.
\newblock \emph{Simon \& Schuster}
\bibAnnoteFile{Ledoux}

\bibitem[{Lewis(2000)}]{lewis2000self}
Lewis, M. (2000).
\newblock Self-conscious emotions.
\newblock \emph{Emotions} , 742
\bibAnnoteFile{lewis2000self}

\bibitem[{Lewis and Ramsay(1995)}]{lewis1995}
Lewis, M. and Ramsay, D.~S. (1995).
\newblock Developmental changes in infants' responses to stress.
\newblock \emph{Child Development} 66(3), 657--670
\bibAnnoteFile{lewis1995}

\bibitem[{Lieberman et~al.(2007)Lieberman, Eisenberger, Crockett, Tom, Pfeifer,
  and Way}]{Lieberman2007PuttingFI}
Lieberman, M.~D., Eisenberger, N.~I., Crockett, M.~J., Tom, S.~M., Pfeifer,
  J.~H., and Way, B.~M. (2007).
\newblock Putting feelings into words: affect labeling disrupts amygdala
  activity in response to affective stimuli.
\newblock \emph{Psychological science} 18 5, 421--8
\bibAnnoteFile{Lieberman2007PuttingFI}

\bibitem[{Lillicrap et~al.(2015)Lillicrap, Hunt, Pritzel, Heess, Erez, Tassa
  et~al.}]{DDPG}
Lillicrap, T.~P., Hunt, J.~J., Pritzel, A., Heess, N., Erez, T., Tassa, Y.,
  et~al. (2015).
\newblock Continuous control with deep reinforcement learning.
\newblock \emph{arXiv preprint arXiv:1509.02971}
\bibAnnoteFile{DDPG}

\bibitem[{Lim and Okuno(2015)}]{lim2015recipe}
Lim, A. and Okuno, H.~G. (2015).
\newblock A recipe for empathy.
\newblock \emph{International Journal of Social Robotics} 7, 35--49
\bibAnnoteFile{lim2015recipe}

\bibitem[{Marchewka et~al.(2014)Marchewka, {\.Z}urawski, Jednor{\'o}g, and
  Grabowska}]{marchewka2014nencki}
Marchewka, A., {\.Z}urawski, {\L}., Jednor{\'o}g, K., and Grabowska, A. (2014).
\newblock The nencki affective picture system (naps): Introduction to a novel,
  standardized, wide-range, high-quality, realistic picture database.
\newblock \emph{Behavior research methods} 46, 596--610
\bibAnnoteFile{marchewka2014nencki}

\bibitem[{Masuyama and Loo(2015)}]{RoboticEmotional}
Masuyama, N. and Loo, C.~K. (2015).
\newblock Robotic emotional model with personality factors based on
  pleasant-arousal scaling model.
\newblock \emph{In Robot and Human Interactive Communication (RO-MAN), 2015
  24th IEEE International Symposium on. IEEE} , 19--24
\bibAnnoteFile{RoboticEmotional}

\bibitem[{Mendoza and Foundas(2007)}]{mendoza2007clinical}
Mendoza, J. and Foundas, A. (2007).
\newblock \emph{Clinical Neuroanatomy: A Neurobehavioral Approach} (Springer
  New York)
\bibAnnoteFile{mendoza2007clinical}

\bibitem[{Miwa et~al.(2001)Miwa, Umetsu, Takanishi, and Takanobu}]{miwa01}
Miwa, H., Umetsu, T., Takanishi, A., and Takanobu, H. (2001).
\newblock Robot personality based on the equations of emotion defined in the 3d
  mental space.
\newblock In \emph{Proceedings 2001 ICRA. IEEE International Conference on
  Robotics and Automation}. vol.~3, 2602--2607.
\newblock \doi{10.1109/ROBOT.2001.933015}
\bibAnnoteFile{miwa01}

\bibitem[{Mnih et~al.(2014)Mnih, Heess, Graves, and Kavukcuoglu}]{RAM2014}
Mnih, V., Heess, N., Graves, A., and Kavukcuoglu, K. (2014).
\newblock Recurrent models of visual attention.
\newblock In \emph{NIPS}
\bibAnnoteFile{RAM2014}

\bibitem[{Mnih et~al.(2015)Mnih, Kavukcuoglu, Silver, Rusu, Veness, Bellemare
  et~al.}]{mnih2015human}
Mnih, V., Kavukcuoglu, K., Silver, D., Rusu, A.~A., Veness, J., Bellemare,
  M.~G., et~al. (2015).
\newblock Human-level control through deep reinforcement learning.
\newblock \emph{Nature} 518, 529--533
\bibAnnoteFile{mnih2015human}

\bibitem[{Moerland et~al.(2017)Moerland, Broekens, and
  Jonker}]{moerland2017emotion}
Moerland, T.~M., Broekens, J., and Jonker, C.~M. (2017).
\newblock Emotion in reinforcement learning agents and robots: A survey.
\newblock \emph{arXiv preprint arXiv:1705.05172}
\bibAnnoteFile{moerland2017emotion}

\bibitem[{Murray et~al.(2016)Murray, De~Pascalis, Bozicevic, Hawkins, Sclafani,
  and Ferrari}]{Murray16}
Murray, L., De~Pascalis, L., Bozicevic, L., Hawkins, L., Sclafani, V., and
  Ferrari, P.~F. (2016).
\newblock The functional architecture of mother-infant communication, and the
  development of infant social expressiveness in the first two months.
\newblock \emph{Scientific Reports} 6:39019, 1--9
\bibAnnoteFile{Murray16}

\bibitem[{Myers(2010)}]{mayer2010}
Myers, D. (2010).
\newblock \emph{Psychology} (Worth publishers)
\bibAnnoteFile{mayer2010}

\bibitem[{Nishihara et~al.(2017)Nishihara, Nakamura, and Nagai}]{nagai2017}
Nishihara, J., Nakamura, T., and Nagai, T. (2017).
\newblock Online algorithm for robots to learn object concepts and language
  model.
\newblock \emph{IEEE Transactions on Cognitive and Developmental Systems} 9,
  255--268.
\newblock \doi{10.1109/TCDS.2016.2552579}
\bibAnnoteFile{nagai2017}

\bibitem[{Ortony et~al.(1988)Ortony, Clore, and Collins}]{OCCbook}
Ortony, A., Clore, G., and Collins, A. (1988).
\newblock \emph{The Cognitive Structure of Emotion}, vol.~18
\bibAnnoteFile{OCCbook}

\bibitem[{Papez(1937)}]{papez1937}
Papez, J. (1937).
\newblock A proposed mechanism of emotion.
\newblock \emph{Arch Neurol Psychiatry} 79, 217--224
\bibAnnoteFile{papez1937}

\bibitem[{Picard(1997)}]{AffectiveComputing}
Picard, R. (1997).
\newblock Affective computing.
\newblock \emph{MIT Press. Cambridge}
\bibAnnoteFile{AffectiveComputing}

\bibitem[{Plutchik(1980)}]{plutchik1980emotion}
Plutchik, R. (1980).
\newblock \emph{Emotion: A Psychoevolutionary Synthesis} (Harper and Row)
\bibAnnoteFile{plutchik1980emotion}

\bibitem[{Plutchik(1982)}]{Plutchik}
Plutchik, R. (1982).
\newblock A psychoevolutionary theory of emotions.
\newblock \emph{Social Science Information} , 529--553
\bibAnnoteFile{Plutchik}

\bibitem[{Russell(1980)}]{Russell}
Russell, J. (1980).
\newblock A circumplex model of affect 39, 1161--1178
\bibAnnoteFile{Russell}

\bibitem[{Ruvolo et~al.(2015)Ruvolo, Messinger, and Movellan}]{Ruvolo15}
Ruvolo, P., Messinger, D., and Movellan, J. (2015).
\newblock Infants time their smiles to make their moms smile.
\newblock \emph{PLOS ONE} 10, 1--10.
\newblock \doi{10.1371/journal.pone.0136492}
\bibAnnoteFile{Ruvolo15}

\bibitem[{Schachter and Singer(1962)}]{Schachter}
Schachter, S. and Singer, J. (1962).
\newblock Cognitive, social, and physiological determinants of emotional state.
\newblock \emph{Psychological Review} 69(5), 379--399
\bibAnnoteFile{Schachter}

\bibitem[{Schlosberg(1954)}]{Schlosberg1954}
Schlosberg, H. (1954).
\newblock Three dimensions of emotion.
\newblock \emph{Psychological Review} 61(2), 81--88
\bibAnnoteFile{Schlosberg1954}

\bibitem[{Seth and Friston(2016)}]{Seth20160007}
Seth, A.~K. and Friston, K.~J. (2016).
\newblock Active interoceptive inference and the emotional brain.
\newblock \emph{Philosophical Transactions of the Royal Society of London B:
  Biological Sciences} 371.
\newblock \doi{10.1098/rstb.2016.0007}
\bibAnnoteFile{Seth20160007}

\bibitem[{Terasawa et~al.(2013)Terasawa, Fukushima, and
  Umeda}]{Terasawa2013humanbrainmapping}
Terasawa, Y., Fukushima, H., and Umeda, S. (2013).
\newblock How does interoceptive awareness interact with the subjective
  experience of emotion? an fmri study.
\newblock \emph{Human Brain Mapping} 34, 598--612.
\newblock \doi{10.1002/hbm.21458}
\bibAnnoteFile{Terasawa2013humanbrainmapping}

\bibitem[{Winnicott(1960)}]{Winnicott60}
Winnicott, D. (1960).
\newblock The theory of the parent--infant relationship.
\newblock \emph{International Journal of Psychoanalysis} 41, 585--595
\bibAnnoteFile{Winnicott60}

\bibitem[{Woo et~al.(2015)Woo, Botzheim, and Kubota}]{Verbalconversation}
Woo, J., Botzheim, J., and Kubota, N. (2015).
\newblock Verbal conversation system for a socially embedded robot partner
  using emotional model.
\newblock \emph{In Robot and Human Interactive Communication (RO-MAN), 2015
  24th IEEE International Symposium on. IEEE} , 37--42
\bibAnnoteFile{Verbalconversation}

\bibitem[{Xingjian et~al.(2015)Xingjian, Chen, Wang, Yeung, Wong, and
  Woo}]{xingjian2015convolutional}
Xingjian, S., Chen, Z., Wang, H., Yeung, D.-Y., Wong, W.-K., and Woo, W.-c.
  (2015).
\newblock Convolutional lstm network: A machine learning approach for
  precipitation nowcasting.
\newblock In \emph{Advances in neural information processing systems}. 802--810
\bibAnnoteFile{xingjian2015convolutional}

\bibitem[{Yakovlev(1948)}]{YAKOVLEV1948}
Yakovlev, P. (1948).
\newblock Motility, behavior and the brain. stereodynamic organization and
  neural co-ordinates of behavior.
\newblock \emph{J. Nerv. Ment. Dis.} 107, 313--335
\bibAnnoteFile{YAKOVLEV1948}

\bibitem[{Yamawaki(2010)}]{color2010}
Yamawaki, K. (2010).
\newblock \emph{A book that understands all of color psychology} (Natsumesha
  CO.,LTD.)
\bibAnnoteFile{color2010}

\end{thebibliography}
